\documentclass[lettersize,journal]{IEEEtran}
\usepackage{amsmath,amsfonts}
\usepackage{algorithmic}
\usepackage{algorithm}
\usepackage{array}
\usepackage{textcomp}
\usepackage{stfloats}
\usepackage{url}
\usepackage{verbatim}
\usepackage{graphicx}
\usepackage{cite}
\usepackage{color}
\usepackage{colortbl}
\usepackage{soul}
\usepackage{multirow}
\usepackage{multicol}
\usepackage{makecell}
\usepackage{booktabs}
\usepackage{hyperref}
\usepackage{booktabs}
\usepackage{amssymb}
\usepackage{bbding}
\usepackage{pifont}
\usepackage{wasysym}
\usepackage{subfigure}
\usepackage{amsfonts}
\usepackage{pifont}
\usepackage{enumitem}
\usepackage[table]{xcolor}

\usepackage{amsmath}
 
\newcolumntype{P}[1]{>{\centering\arraybackslash}p{#1}}
\newcolumntype{C}[1]{>{\centering\arraybackslash}m{#1}}

\def \textHT [#1]{\color{red}$\mathbf{#1}$\color{black}}
\def \textLT [#1]{\color{blue}$\mathbf{#1}$\color{black}}
\newcommand{\rt}{\textcolor[rgb]{1,0,0}}

\begin{document}


\title{General Place Recognition Survey: Towards Real-World Autonomy
}

\author{Peng Yin\textsuperscript{1,$^\dagger$, *},
    Jianhao Jiao\textsuperscript{2,$^\dagger$},
    Shiqi Zhao\textsuperscript{1},
    Lingyun Xu\textsuperscript{3}, \\
    Guoquan Huang\textsuperscript{4},
    Howie Choset\textsuperscript{3},
    Sebastian Scherer\textsuperscript{3},
    and Jianda Han\textsuperscript{5}
    \thanks{Peng Yin and Shiqi Zhao are with the Department of Mechanical Engineering, City University of Hong Kong, Hong Kong 518057, China. {(pengyin@andrew.cmu.edu, ryanzhao9459@gmail.com)}.}
    \thanks{Jianhao Jiao is with the Department of Computer Science, University College London, Gower Street, WC1E 6BT, London, UK. {(ucacjji@ucl.ac.uk)}.}
    \thanks{Guoquan Huang is with the Robot Perception and Navigation Group, University of Delaware, Newark, DE 19716 USA. {(ghuang@udel.edu)}.}
    \thanks{Linyun Xu, Howie Choset and Sebastian Scherer are with the Robotics Institute, Carnegie Mellon University, Pittsburgh, PA 15213, USA. {(xulinyun2021@gmail.com, (choset, basti)@andrew.cmu.edu)}.}
    \thanks{Jiandan Han is with Nankai University, Tianjin, 300071, China. {(hanjianda@nankai.edu.cn)}.}
    \thanks{*Corresponding author: Peng Yin (pengyin@andrew.cmu.edu)}
    \thanks{$^\dagger$Peng Yin and Jianhao Jiao Contributed Equally.}
}


\maketitle

\begin{abstract}
    In the realm of robotics, the quest for achieving real-world autonomy, capable of executing large-scale and long-term operations, has positioned place recognition (PR) as a cornerstone technology.
    Despite the PR community's remarkable strides over the past two decades, garnering attention from fields like computer vision and robotics, the development of PR methods that sufficiently support real-world robotic systems remains a challenge.
    This paper aims to bridge this gap by highlighting the crucial role of PR within the framework of Simultaneous Localization and Mapping (SLAM) 2.0. This new phase in robotic navigation calls for scalable, adaptable, and efficient PR solutions by integrating advanced artificial intelligence (AI) technologies.
    For this goal, we provide a comprehensive review of the current state-of-the-art (SOTA) advancements in PR, alongside the remaining challenges, and underscore its broad applications in robotics.

    This paper begins with an exploration of PR's formulation and key research challenges.
    We extensively review literature, focusing on related methods on place representation and solutions to various PR challenges.
    Applications showcasing PR's potential in robotics, key PR datasets, and open-source libraries are discussed.
    We conclude with a discussion on PR's future directions and provide a summary of the literature covered at:
    \href{https://github.com/MetaSLAM/GPRS}{https://github.com/MetaSLAM/GPRS}.


\end{abstract}


\begingroup
\let\clearpage\relax
\section{Introduction}
\label{sec:introduction}

\begin{figure}[t]
    \begin{center}
        \includegraphics[width=0.97\linewidth]{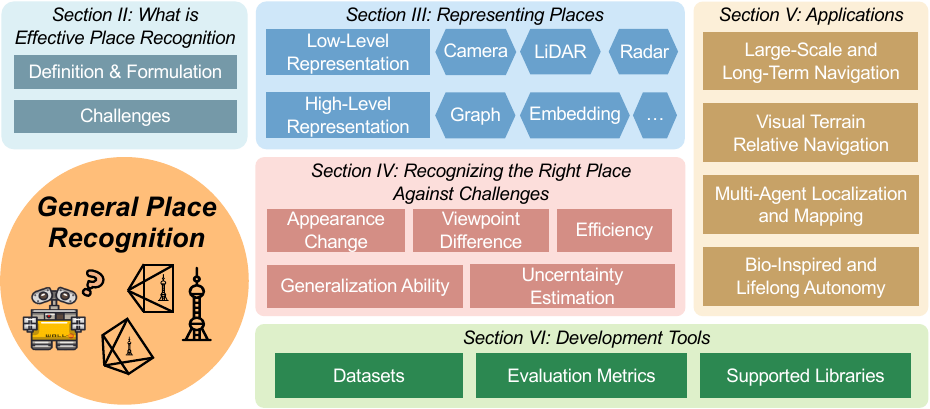}
    \end{center}
    \caption{
        \textbf{Structure of our General Place Recognition (PR) Survey.}
        PR is the ability to recognize visited areas under different environmental conditions and viewpoint differences.
        This survey is structured as follows:
        Section \ref{sec:definition_challenges} defines the problem of position-based PR and introduces the significant challenges.
        Section \ref{sec:representation} investigates methods in place representation.
        Section \ref{sec:solution} and Section \ref{sec:application} provide the solutions for the current four major challenges and potential applications, respectively,
        Finally, Section \ref{sec:data_eval} introduces the current datasets, metrics, and related supported libraries for PR research.}
    \label{fig:idea}
    \vspace{-0.5cm}
\end{figure}

\begin{figure*}[t]
  \begin{center}
    \includegraphics[width=0.90\linewidth]{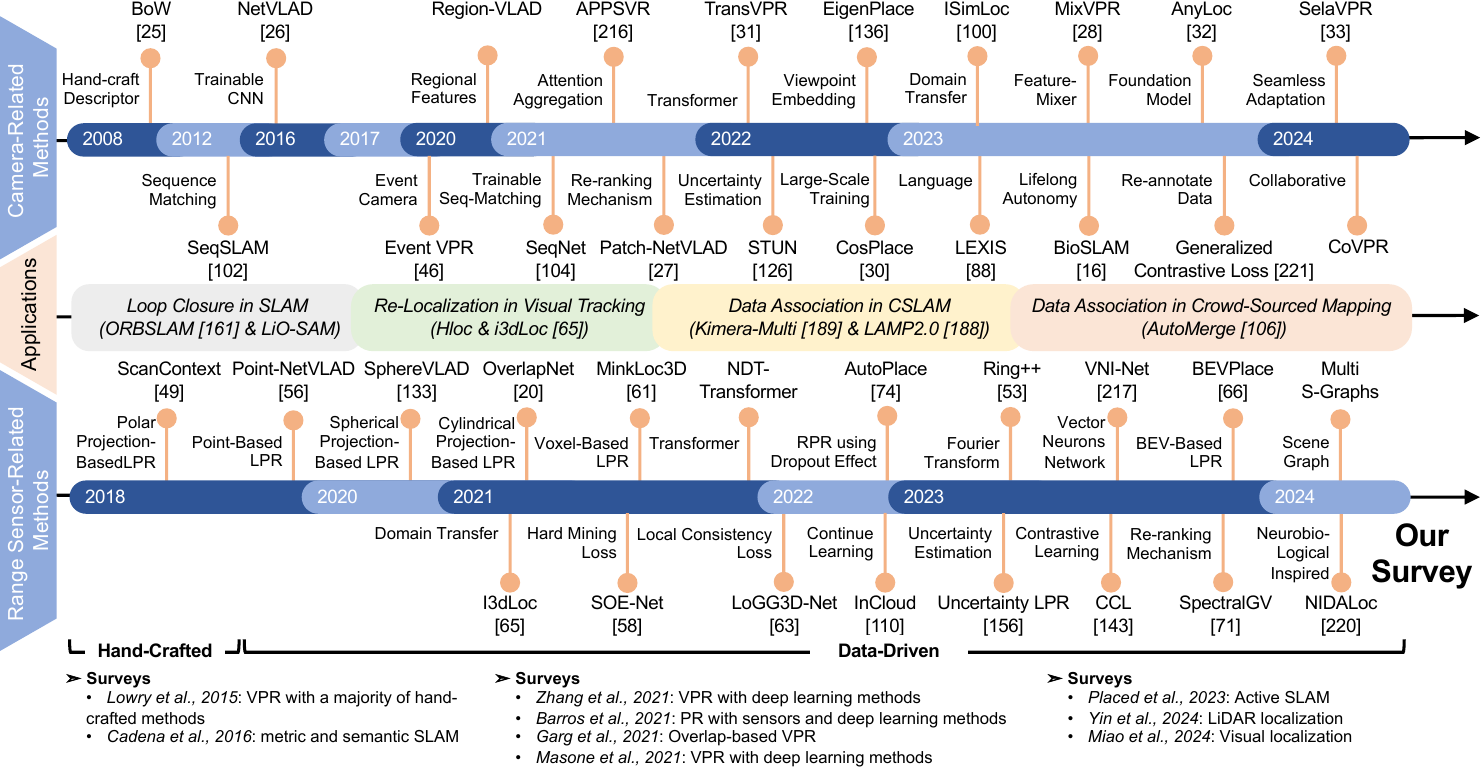}
  \end{center}
  \vspace{-0.3cm}
  \caption{This timeline maps the evolution of PR from handcrafted to data-driven methods, analyzing key techniques, surveys, and applications.
    Our survey emerges at an opportune moment, given major events across multiple fields including embodied AI, reconstruction, and collaborative perception. Citations for select works are omitted due to limited page; readers may search referenced titles online.}
  \label{fig:timeline_pr}
  \vspace{-0.5cm}
\end{figure*}


\subsection{Background}

\IEEEPARstart{I}{n} recent decades, mobile robot systems have gained significant interest for their roles in diverse applications, such as autonomous driving, last-mile delivery, search-and-rescue operations, and warehouse logistics.
These robots are increasingly woven into the fabric of our daily routines, facing growing demands for navigating complex environments.
This evolution prompts a critical inquiry: \textit{How can robots achieve lifelong autonomy with the capability of zero-shot or few-shot transferring to new environments and tasks?}

As the fundamental module in navigation, SLAM enables a robot to estimate its ego-motion while simultaneously constructing a map of its environment.
Lifelong navigation, the capability for robots to autonomously adapt to ever-changing environments with increasing experience, drives SLAM systems beyond accuracy-focused metrics, demanding solutions for long-term localization, dynamic mapping, and self-evolution.
This signals the dawn of the \textbf{SLAM}$2.0$ era: diverging from traditional SLAM frameworks \cite{SURVEY:VPR},
SLAM$2.0$ prioritizes the synergy of advanced techniques such as machine learning-driven perception and semantic scene abstraction to enable scalable and efficient robotic autonomy in dynamic, open-world environments.

As the core of advancing to SLAM$2.0$, PR is now becoming more essential than ever, which enables robots to identify previously visited areas despite changes in environmental conditions and viewpoints.
Essentially, PR's ability hinges on creating short-term or long-term association between current observations and a robot's internal ``memory'' of environments.
In visual SLAM (VSLAM), memory typically refers to a map (or called database in some context) that consists of visual information.
For decision-making, PR enables robots to associate and apply past experiences to the current situation.
Thus, PR's role extends beyond merely loop closure detection (LCD) in SLAM.
Its applications now span failure recovery, global localization, multi-agent coordination, and more.

PR has emerged as a cornerstone capability for robotic autonomy, driven by three key indicators of its growing importance:
(1) Over $\mathbf{3500}$ papers have been published on PR methodologies to date;
(2) the transition from controlled indoor settings to unstructured outdoor environments, with challenges like perceptual aliasing and dynamic conditions, has necessitated robust, scalable PR solutions;
and
(3) the field has seen a surge in organized competitions and workshops,
including
the \textit{CVPR 2020 Long-term Visual PR (VPR) Challenge},
the \textit{ICRA 2022 General PR competition for \href{https://www.aicrowd.com/challenges/icra2022-general-place-recognition-city-scale-ugv-localization}{City-scale UGV Localization} and \href{https://www.aicrowd.com/challenges/icra2022-general-place-recognition-visual-terrain-relative-navigation}{Visual Terrain Relative Navigation} (VTRN)},
and
the \textit{IROS 2023 Closing the Loop on Localization Workshop}.
These converging factors position PR as a critical frontier in autonomous systems, catalyzing decades of methodological innovation spanning geometric, semantic, and learning-based paradigms.



\subsection{Summary of Previous Surveys}
\label{sec:intro_summary_survey}
The evolution of PR is visualized as the timeline in Fig. \ref{fig:timeline_pr}.
A comprehensive historical analysis up to the year $2015$ on VPR is provided by Lowry \textit{et al.} \cite{lowry2015visual}.
This seminal survey articulates the VPR challenge, delineates the core components of a PR system, and reviews major solutions to tackle appearance changes.
It distills the essence of a PR system into three fundamental modules which still serves as the foundations of modern solutions:
(1) an image processing module for visual input abstraction,
(2) a map to represent how a robot stores its memory of the environment, and
(3) a belief generation module to evaluate the likelihood of the robot being in a previously encountered or new location.

In the past decade, the evolution of mainstream PR methods has been transitioned from handcrafted descriptors to data-driven pipelines, as comprehensively discussed in prominent studies \cite{zhang2021visual,barros2021place,garg2021your}.
However, as Zaffar \textit{et al.} \cite{zaffar2021vpr} observed, the PR community has developed increasing fragmentation in performance benchmarking. This disparity complicates direct method comparisons due to inconsistent evaluation metrics and dataset implementations across studies. To resolve these challenges, their work established an open-source standardized evaluation framework specifically focused on VPR \cite{zaffar2021vpr}.
Other surveys like \cite{miao2023survey} (vision-based) and \cite{yin2024survey} (LiDAR-based) exclusively address VPR and LPR respectively.
While these works review pose estimation techniques for fine-grained metric localization, such methods fall outside our core scope.


\subsection{Contributions and Paper Organization}
Recent progress in autonomous systems and machine perception has driven dynamic evolution in PR research.
However, existing literature lacks a survey that thoroughly explores the diverse aspects, challenges, and deployment potentials for embodied AI applications.
Our work addresses this gap by introducing a ``General PR'' (GPR) framework, which extends beyond the scope of VPR and LPR to encompass a wider spectrum of topics.
GPR emphasizes the utilization of multi-model information such as visual, geometric, and textual inputs to establish robust environmental embeddings.
This systematic review is particularly pertinent given transformative developments in three key areas:
(1) Foundational language models \cite{achiam2023gpt,bi2024deepseek} demonstrating unprecedented reasoning capabilities;
(2) neural scene representation frameworks \cite{mildenhall2021nerf} enabling photorealistic environment reconstruction; and
(3) expanding real-world robot deployment across industrial and service sectors.
The paper's organizational architecture, visualized in Fig. \ref{fig:idea}, proceeds as follows:
\begin{itemize}[leftmargin=0.5cm]
    \item Section \ref{sec:definition_challenges} details two widely accepted definitions of PR: the position-based and the overlap-based definitions. It then offers a more precise formulation about ``effective PR'' and highlights the key challenges involved.
    \item Section \ref{sec:representation} reviews existing representation approaches in PR, covering the core solutions prevalent in the field. Intuitively, PR extends beyond mere image-based approaches, encompassing a variety of solutions. At a low level, detailed in Section \ref{sec:representation_low_level}, a ``place'' can be captured through sensors, such as cameras, LiDARs and Radars. It's generally expected that identical locations will produce similar sensor data. On a more abstract and higher level, as discussed in Section \ref{sec:reprsentation_high_level}, a ``place'' may also be represented through compact data forms, like scene graphs, implicit embeddings, and Gaussian Splatting (GS) maps.
    \item Section \ref{sec:solution} delves into the primary challenges faced by PR, exploring how contemporary solutions are tailored to achieve key attributes such as invariance to conditions and viewpoints, great generalization ability, high efficiency, and uncertainty awareness.
    \item Section \ref{sec:application} concentrates on the deployment of PR techniques for achieving real-world autonomy. It highlights opportunities from these aspects:
          large-scale and long-term navigation (Section \ref{sec:app_longterm}), visual terrain relative navigation (VTRN) (Section \ref{sec:vtrn}),  multi-agent localization and mapping (Section \ref{sec:app_multi}), and lifelong navigation (Section \ref{sec:lifelong_autonomy}).
          We posit that PR is poised to become a cornerstone in modern robotics, with its applications and related research beyond the realms of SLAM.
    \item Section \ref{sec:data_eval} reviews the leading datasets and benchmarks in the field of PR. It introduces a new perspective on property analysis that complements to primary metrics for quality assessment.
    \item Section VII provides a thorough conclusion of this survey and outlines potential directions for future research.
\end{itemize}

\section{Formulation of Effective Place Recognition and Challenges}
\label{sec:definition_challenges}

Before exploring specific solutions in PR, it is crucial to address two basic questions:
(1) \textit{What is effective PR?} and
(2) \textit{What are the primary challenges encountered in PR?}




\subsection{What is Effective Place Recognition?}
\label{sec:def_definition}

\subsubsection{Existing Definitions}
Two principal paradigms dominate PR definitions: position-based and overlap-based.
Fig. \ref{fig:definition_overlap_positon_pr} explains their differences in place judgement with an example.
Originating from O'Keefe's discovery of hippocampal ``place cells'' \cite{place_cell},
\textbf{position-based PR} \cite{lowry2015visual} evaluates whether a robot revisits a geographic location (point or region) despite environmental or viewpoint changes. The core challenge lies in robustly associating observations with spatial proximity. Conversely, \textbf{overlap-based PR} \cite{garg2021your} defines place equivalence as visual overlap in the sensor's field-of-view (FoV), irrespective of geographic distance, aligning with image retrieval \cite{google_landmark}.

While overlap-based PR mirrors content-based image search, its utility in robotic SLAM and navigation remains ambiguous. For instance, Fig. \ref{fig:definition_overlap_positon_pr} illustrates two images observing the same landmark (e.g., a building) from distinct viewpoints: despite visual overlap, inferring their relative positions is non-trivial. Position-based PR better supports tasks like global localization \cite{yin2024survey}, which prioritize coarse pose estimation. However, neither paradigm universally addresses all PR challenges. Thus, we emphasize \textbf{effective PR} as the intersection of two criteria:
(1) Geographic proximity: places share meaningful spatial adjacency, and
(2) visual consistency: observations exhibit measurable scene overlap or descriptor similarity.

\subsubsection{Formulation}
Effective PR assumes a valid database image for the query image capture at place $Q$ must satisfy:
\begin{itemize}[leftmargin=0.5cm]
    \item Geometric constraint: The candidate place $P \in \mathcal{P}$ is within a threshold distance $\delta$ from $Q$:
          \begin{equation}
              d(P, Q) < \delta,
          \end{equation}
          where $d(\cdot,\cdot)$ measures translational or rotational distance.
    \item Visual constraint: Their global descriptors $\mathbf{g}_P$ and $\mathbf{g}_Q$ exhibit high similarity with $\epsilon$ as a similarity threshold:
          \begin{equation}
              \|\mathbf{g}_P - \mathbf{g}_Q\| < \epsilon.
          \end{equation}
\end{itemize}

This formulation excludes ambiguous cases (e.g., places on opposite sides of a wall or distant viewpoints observing the same landmark). A PR method succeeds only if it retrieves places satisfying both conditions, ensuring robustness to perceptual aliasing while maintaining spatial relevance.

\begin{figure}[t]
  \begin{center}
    \includegraphics[width=0.77\linewidth]{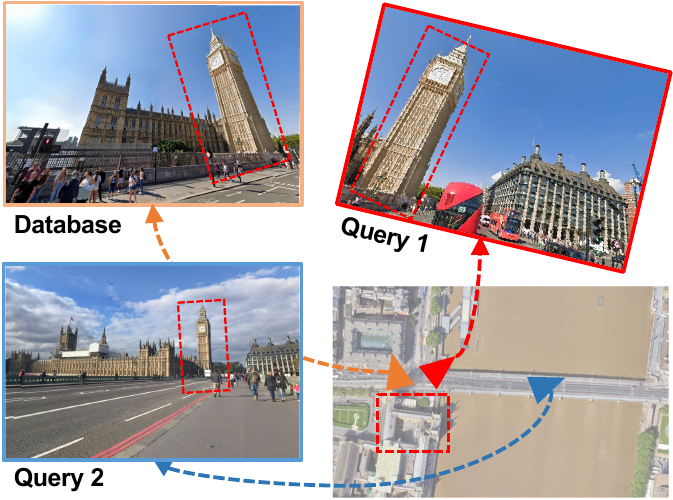}
  \end{center}
  \vspace{-0.3cm}
  \caption{In position-based PR, the focus is on identifying if a query image stays at the same location as a database image.
    For instance, in the provided images, position-based PR would recognize the Query $1$ image as matching the database image,
    but reject the Query $2$ that is taken from a geographically distant location and considered as a different place.
    Overlap-based PR, however, would classify both query images as the same place since they share visual overlap with the database, denoted by the red box. But the Query $2$ offers limited utility for downstream navigation tasks.}
  \label{fig:definition_overlap_positon_pr}
  \vspace{-0.5cm}
\end{figure}

\begin{figure*}[t]
    \centering
    \subfigure[Appearance Change]{
        \label{fig:challenge_appearance_change}
        \centering
        \includegraphics[width=0.135\textwidth]{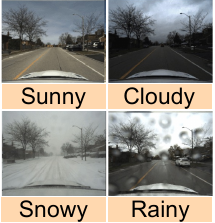}
    }
    \hspace{-0.4cm}
    \subfigure[Viewpoint Difference]{
        \label{fig:challenge_viewpoint_difference}
        \centering
        \includegraphics[width=0.28\textwidth]{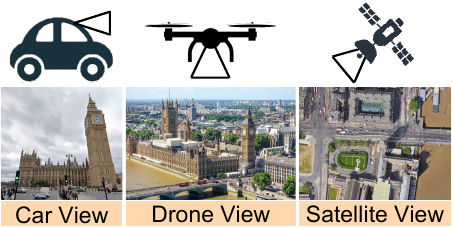}
    }
    \hspace{-0.4cm}
    \subfigure[Generalization Ability]{
        \label{fig:challenge_generalization_ability}
        \centering
        \includegraphics[width=0.230\textwidth]{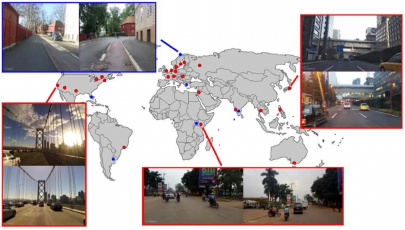}
    }
    \hspace{-0.45cm}
    \subfigure[Efficiency]{
        \label{fig:challenge_efficiency}
        \centering
        \includegraphics[width=0.17\textwidth]{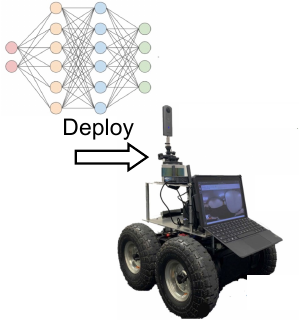}
    }
    \hspace{-0.45cm}
    \subfigure[Uncertainty Estimation]{
        \label{fig:challenge_uncertainty_estimation}
        \centering
        \includegraphics[width=0.16\textwidth]{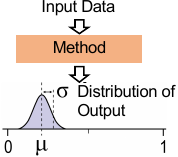}
    }
    \caption{Challenges in real-world PR.
        In real-world navigation tasks, robots may encounter the following challenges:
        (a) changing visual appearances due to temporal variations (lighting, seasons) \cite{burnett2023boreas},
        (b) diverse viewpoint differences for the same areas,
        (c) visiting new unknown areas \cite{warburg2020mapillary},
        (d) impacts to efficiency when deployed on real-world robots \cite{yin2023bioslam},
        and
        (e) uncertainty estimation of data and modal \cite{dolezal2022uncertainty}.}
    \label{fig:pr_challenge}
    \vspace{-0.35cm}
\end{figure*}

\subsection{Challenges}
\label{sec:def_challenge}

As summarized in Fig. \ref{fig:pr_challenge}, solving the position-based PR problem for real-world robot autonomy requires addressing five specific categories of practical challenges:
\begin{enumerate}[leftmargin=0.5cm]
    \item \textbf{Appearance Change}: Compared to short-term navigation, long-term operation may contain appearance changes under different illumination conditions or structural changes (\textit{i.e.,} parking lot and construction sites), which will introduce further localization failures.
    \item \textbf{Viewpoint Difference}: This issue arises from the variations in how the environment is captured by sensors, which can be influenced by the sensor's viewpoint, position, and intrinsic properties.
          For example, a building's frontal view exposes its full shape, whereas a top-down perspective highlights the layout and roof design. Such viewpoint variations are especially noticeable when a robot revisits a location from a different angle or altitude. This challenge is ubiquitous across sensor types, necessitating PR systems to incorporate robust modules for feature extraction and matching to accommodate these perspective shifts.
    \item \textbf{Generalization Ability}: For lifelong navigation, the vast complexity of environments makes generalizing to unseen areas a crucial challenge. PR methods must enable online learning to adapt over time, an essential requirement for missions like space exploration, where robots conduct long-term investigations on other planets with minimal human supervision. Robots need the ability to continuously learn and adjust to new environments.
    \item \textbf{Efficiency on Resource-Constrained Platforms}: Deploying PR algorithms, particularly those requiring online fine-tuning, on resource-constrained robotic platforms is a major challenge, especially for aerial robots. Effective algorithms must strike a balance between accuracy and computational efficiency. This is essential for both single and multi-robot systems, which often encounter bandwidth limitations and potential communication disruptions.
    \item \textbf{Uncertainty Estimation}: Generating a belief distribution for assessing likelihood or confidence, identifying out-of-distribution data, and evaluating PR algorithm reliability is crucial for downstream navigation tasks like PGO and mapping. But accurately estimating and qualifying the likelihood function is challenging.
\end{enumerate}

Building from above formulation and challenges, the following sections will delve into associate studies.

\section{Describing Places: Representation}
\label{sec:representation}


The foundation of PR lies in how a place is represented as a map and then compared with incoming sensor data. Based on existing representation formats, we categorize these into low-level, sensor-specific representations and high-level, sensor-agnostic representations.


\subsection{Low-Level Representations}
\label{sec:representation_low_level}

This approach represents a place as a database of primitive representations: images, point clouds, or features extracted from raw data with methods tailored to a specific sensor.

\begin{figure*}[t]
    \begin{center}
        \includegraphics[width=0.85\linewidth]{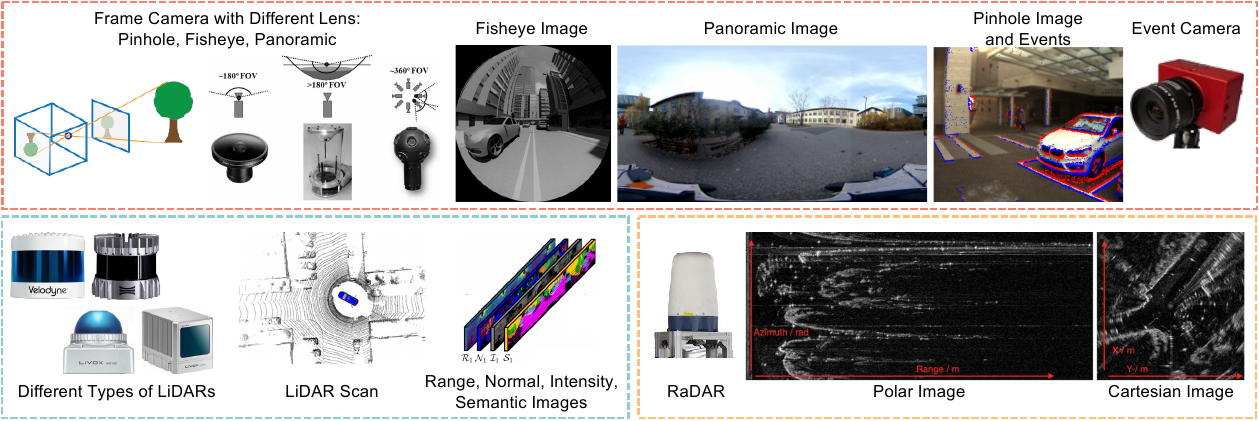}
    \end{center}
    \vspace{-0.3cm}
    \caption{Diverse Sensor Modalities and Observation Properties.
        The top box contains various camera setups with different lens \cite{scaramuzza2014omnidirectional} and imaging sensors \cite{jiao2022fusionportable}. The bottom-left box shows major LiDAR types, point cloud, and multi-channel images using the cylinder projection from the point cloud \cite{chen2021overlapnet}. The bottom-right box shows a typical RaDAR and data represented in polar and cartesian images \cite{hong2022radarslam}.}
    \label{fig:sensor}
    \vspace{-0.5cm}
\end{figure*}

\subsubsection{Sensor Selection Criteria}
\label{sec:representation_sensor}
Cameras, LiDAR (Light Detection and Ranging), and Radar (Radio Detection and Ranging) are typical sensors in PR, as illustrated in Fig. \ref{fig:sensor}.
Important selection criteria include FoV, information density, and robustness under various conditions. Frame cameras provide high-resolution images, while event cameras perform better in low-light and reduce motion blur.
LiDARs generate precise 3D point clouds but have low resolution.
Radars offer long-range capabilities and excel in poor weather, also measuring relative velocity via the Doppler effect.
Choosing the proper sensor or sensor combinations for PR depends on specific requirements for precision, range, and environmental suitability.

\subsubsection{Camera-Related Approaches}
\label{sec:representation_image_based_approach}

VPR is the most thoroughly investigated problem.
As highlighted in previous surveys (Section \ref{sec:intro_summary_survey}),
it has been comprehensively explored through both handcrafted and data-driven methodologies.

\textit{\textbf{Handcrafted VPR Methods:}}
Handcrafted representations in VPR are generally divided into local and global feature descriptors.
Local descriptors such as SURF \cite{bay2006surf} identify a set of keypoints and compute corresponding descriptors for local image regions, but requiring extensive matching to compare visual similarity between images.
In contrast, global descriptors, such as Gist \cite{oliva2006building}, CoHOG \cite{Zaffar2020CoHoG}, and BoW \cite{galvez2012bags}, aggregate local features into a unified descriptor, such as a vector or matrix, to capture an image's overall pattern without local matching.
More extensive review on tranditional methods can be referred to this survey paper \cite{lowry2015visual}.
While effective, these handcrafted methods generally underperform compared to data-driven, deep-learning approaches, which offer superior accuracy and robustness.

\textit{\textbf{Data-Driven VPR Methods:}}
Data-driven approaches \cite{relja2016netvlad,hausler2021patch,ali2023mixvpr,radenovic2018fine,berton2022rethinking,wang2022transvpr,keetha2023anyloc,feng2024selavpr}, particularly those utilizing deep neural networks (DNNs), automatically learn features from training data, reducing the need for extensive handcrafted design and domain expertise. A common two-step pipeline is often employed to enhance PR robustness:
(1) initial matching and candidate selection by comparing global descriptors between the query and database data, and
(2) re-ranking by refining matches using local features among selected candidates.

The emergence of CNN provided a new way to generate visual representations which had been proven to be successful in various category-level recognition tasks~\cite{VPR:ALEXNET,zhou2017places}.
The introduction of NetVLAD \cite{relja2016netvlad} represented a major step forward in data-driven VPR, using CNNs to transform images into feature maps, with a differentiable VLAD (Vector of Locally Aggregated Descriptors) pooling layer to create global descriptors. Later methods refined both feature extraction and aggregation, such as Regional Maximum Activations of Convolutions (R-MAC) \cite{tolias2015particular} and Generalized Mean (GeM) \cite{radenovic2018fine,berton2022rethinking} as effective VLAD alternatives. Recent efforts also explored connecting global descriptors with local features.

Rather than focusing solely on enhancing local feature extraction from images, researchers are increasingly emphasizing the relationships between these features, which encapsulate valuable semantic information. Attention mechanisms and transformer architectures address this need by dynamically weighting feature relationships across the entire image, overcoming the constrained receptive field of CNNs while fostering more robust contextual semantics.
Retriever \cite{retriever2022}, using cross-attention, and TransVPR \cite{wang2022transvpr} with multi-scale self-attention. Beyond transformers, MixVPR \cite{ali2023mixvpr}, inspired by isotropic all-MLP architectures like MLP-Mixer \cite{mlpmixer2021}, proposes a novel all-MLP aggregation that performs competitively in one stage. SALAD \cite{salad2024} addresses the assignment in NetVLAD with an optimal transport approach using the Sinkhorn Algorithm \cite{sinkhorn2013}.

Foundation models pre-trained on extensive datasets, like DINOv2 \cite{oquab2023dinov2}, exhibit strong zero- and few-shot generalizability for VPR. Methods such as AnyLoc \cite{keetha2023anyloc} and SelaVPR \cite{feng2024selavpr} leverage DINOv2 to achieve state-of-the-art (SoTA) results. Additionally, PR results generated by global descriptors are still not accurate enough, especially for robot navigation tasks that require high top-$1$ recall. To enhance accuracy, several methods integrate local features. Patch-NetVLAD \cite{hausler2021patch} pioneered re-ranking by generating sub-global descriptors using patches with NetVLAD. R2former \cite{r2former2023} provides an end-to-end approach for computing re-ranking scores.


Various research efforts have broadened scope by integrating diverse cues such as semantic, geometric, event data to boost the place representation.
The utilization of semantics includes the way of filtering specific pixels \cite{piasco2019learning} and adjusting the weight of feature embeddings \cite{peng2021semantic}.
Geometric cues, such as the 3D positions of landmarks, provide complementary structural information to visual descriptors and enhance PR accuracy \cite{oertel2020augmenting}.
Event cameras, with a higher dynamic range than frame cameras, have been also used in VPR to capture texture information in low-light conditions.
Lee \textit{et al.} \cite{lee2021eventvlad} proposed to use event cameras to capture texture information under low-light condition, constructed edge-based images from event data to achieve PR.



\subsubsection{Range Sensor-Related Approaches}
\label{sec:reprsentation_pointcloud_based_approach}
Research on LPR has significantly progressed, driven by the extensive application of LiDARs in autonomous vehicles and surveying fields.
However, LiDAR measurements are predominantly stored as point clouds, which are characterized by their sparsity and lack of orderly structure. These attributes present challenges for traditional 2D convolution operations.
To leverage CNN, LPR solutions employ advanced point cloud learning architectures, including PointNet \cite{qi2017pointnet} and the Minkowski Engine\cite{choy20194d}.
Radar-based PR (RPR) research, though less mature, is growing, with efforts concentrating on enhancing Radar perception for all-weather functionality. The forthcoming sections will highlight diverse representation techniques in LPR and then introduce initial progress in RPR research.

\textit{\textbf{Handcraft LPR Methods:}}
\label{sec:representation_lidar_based_approach}
Early approaches such as ScanContext \cite{Kim2018scancontext} and ScanContext++ \cite{kim2021scan} encode LiDAR point clouds into bird's-eye-view (BEV) images, where pixel intensities represent height information.
Building on this foundation, Wang \textit{et al.} \cite{wang2020lidar} enhanced rotation invariance in LiDAR PR through LiDAR IRIS, leveraging LoG-Gabor filters for improved feature extraction.
A work extended ScanContext for map matching with OpenStreetMap data \cite{cho2022openstreetmap}, while Ring++ \cite{2023xuring++} applied Radon and Fourier transforms to BEV images, enhancing feature representation through frequency domain analysis.
Recent innovations such as BTC \cite{2024yuanbtc} adopt geometric descriptors by projecting key points onto planes and deriving triangular features from their spatial relationships.

\textit{\textbf{Data-Driven LPR Methods:}}
The transition to data-driven feature learning observed in VPR has extended to LRP, motivated by neural networks' capacity to learn complex geometric relationships from raw sensor data.
However, implementing conventional architectures like CNNs and Transformers for unstructured 3D point clouds necessitates either specialized network designs (e.g., point-voxel transformers) or geometric preprocessing to reconcile irregular point distributions with structured computational paradigms.

Early advancements in LPR emerged through \textbf{point-based} methods, exemplified by PointNet \cite{qi2017pointnet} and PointSift \cite{jiang2018pointsift}, which process raw point clouds without voxelization.
PointNetVLAD \cite{uy2018pointnetvlad} established a foundational framework by merging PointNet's geometric feature extraction with NetVLAD's descriptor learning.
Subsequent efforts expanded this paradigm: LPD-Net \cite{liu2019lpd} introduced graph-based neighborhood modeling to encode spatial relationships, while SOE-Net \cite{xia2021soe} augmented local features through self-attention mechanisms guided by PointSift-derived orientation embeddings \cite{jiang2018pointsift}.
A critical limitation of PointNet-based approaches, however, lies in their sensitivity to viewpoint rotations, which degrades performance under large orientation changes \cite{liu2019lpd}.
To address this, RPR-Net \cite{fan2022rpr} integrated the rotation-invariant SPRIN backbone \cite{you2021prin}, demonstrating improved robustness.
Despite progress, point-wise methods face scalability constraints due to quadratic complexity growth with point cloud density, limiting their practicality for real-time deployment.

Rather than directly manipulating points within neural networks,
two alternative categories of LPR methods utilizes \textbf{voxelization} \cite{komorowski2021minkloc3d, zywanowski2021minkloc3d, vid2022logg3d} and \textbf{projection-based} techniques \cite{Yin2021spherevlad, yin2021i3dloc, luo2023bevplace}. These methods transform point clouds into 3D voxels and 2D grids respectively, serving as a preparatory phase prior to network input.
Regarding the former category,
MinkLoc3D \cite{komorowski2021minkloc3d} employs sparse 3D convolution for feature extraction.
Its successor, MinkLoc3D-SI \cite{zywanowski2021minkloc3d}, incorporates spherical coordinates and intensity data for each 3D point.
Besides leveraging the sparse convolution, LoGG3D-Net \cite{vid2022logg3d} additionally introduces a local consistency loss that steers the network to consistently learn local features during revisits.

Initially, both point-based and voxel-based methods struggle to handle large viewpoint differences. However, by effectively utilizing point cloud projections, these differences can be more easily mitigated through advanced image processing techniques.
Projection-based methods in LPR vary in approach. Cylindrical projection, which converts point cloud rotations to translations in a 2D image, provides yaw-invariance for convolutional processing. OverlapNet series \cite{chen2021overlapnet, ma2022overlaptransformer} utilizes this method, creating multi-channel images with range, intensity, normals, and semantic information, redefining PR as a classification task based on scan overlap. RINet \cite{li2022rinet} further advances it using semantic and geometric features with attention mechanisms.
Spherical projection, as used in SphereVLAD series \cite{Yin2021spherevlad,zhao2022spherevlad2}, offers 3-DoF rotation invariance, essential for consistent 3D coordinate encoding.
Other projection methods include DiSCO \cite{xu2021disco}, which applies a differentiable ScanContext-like representation using polar projection, and BEVPlace \cite{luo2023bevplace}, which transforms point clouds into BEV images with a rotation-invariant network design.

Re-ranking mechanisms have also been incorporated into LPR.
Unlike Patch-NetVLAD, SpectralGV \cite{spectralgv2023} expands this approach by using Spectral Matching to calculate matching confidence.
TReR \cite{trer2023} introduces a transformer-based re-ranking method that relies solely on global descriptors, bypassing the need for local features.

\textit{\textbf{Data-Driven RPR Methods:}}
\label{sec:representation_radar_based_approach}
RPR techniques predominantly utilize polar and Cartesian images derived from Radar measurements.
Kidnapped Radar \cite{suaftescu2020kidnapped} leverages a CNN backbone to process polar images for feature extraction. AutoPlace \cite{cai2021autoplace} enhances accuracy by employing Doppler measurements to eliminate moving objects and applies a specialized network to encode Radar point clouds, integrating spatial and temporal dimensions, and further refines matches using Radar Cross Section histograms.
mmPlace \cite{meng2024mmplace} designed a rotating single-chip Radar platform to enlarge the FoV.
Additionally, advancements in RPR have been made through exploring sequence matching \cite{gadd2020look},
cross-modal data matching with LiDARs \cite{yin2021radar} and overhead images \cite{tang2020self}, self-supervised fusion \cite{tang2020self}, and data augmentation \cite{gadd2021contrastive} strategies to enhance RPR.


\begin{figure}[t]
    \begin{center}
        \includegraphics[width=0.7\linewidth]{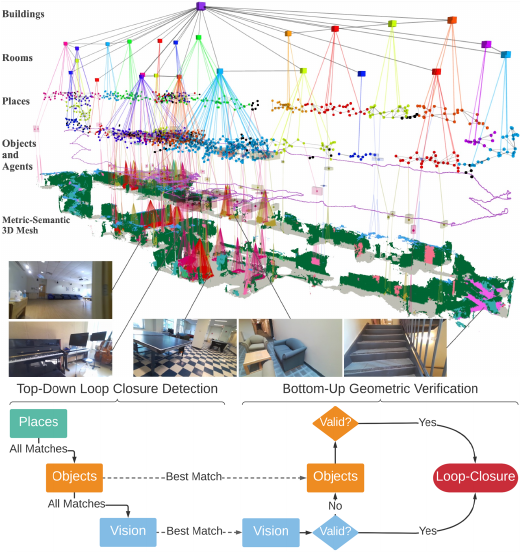}
    \end{center}
    \caption{(Top) A building is abstracted into a hierarchical 3D scene graph by the Hydra system \cite{hughes2022hydra}. (Bottom) Hydra presents a hierarchical PR solution.}
    \label{fig:high_level_representation}
    \vspace{-0.5cm}
\end{figure}

\subsection{High-Level Representations in Various Formats}
\label{sec:reprsentation_high_level}
High-level representations in PR provide a semantic abstraction of environmental structure, diverging from low-level feature-based approaches that rely on raw geometric primitives or sensor-specific measurements.
They capture topological relationships and contextual attributes through structured formats such as graphs or learned embeddings.
Graph-based methods model spatial-semantic interactions (e.g., object adjacency), while neural embeddings project multi-modal inputs (images, point clouds, text) into unified descriptor spaces, enabling cross-domain compatibility through fixed-dimensional vector concatenation.
This abstraction not only decouples PR systems from sensor dependencies but also facilitates hierarchical reasoning about place identity.
In this section, we analyze the design principles of high-level representations, survey their implementation in modern PR frameworks,
and unveil their evolution and role in boosting PR performance.

\subsubsection{Graphs}
\label{sec:representation_graph}
Graphs, including object-based graphs and 3D scene graphs, have recently emerged as a powerful representation of a place.
A graph is a mathematical structure that is used to model pairwise relations between objects.
It consists of vertices (also called nodes) and edges, where the vertices represent the objects and the edges represent the connections or relationships between them.
In the context of PR, these entities can be features, landmarks, or regions, and the edges can denote spatial or topological relations.
Graph representations offer several advantages, including robustness to viewpoint changes, occlusions, and dynamic scenes.
They can also store semantic labels \cite{hughes2022hydra} to enhance PR accuracy.

Recent studies \cite{kong2020semantic,vidanapathirana2021locus, hughes2022hydra, kim2023topological,stumm2013probabilistic,wang2024sglc} have introduced various graph models to depict places and environments.
Co-visibility graphs depict the relationships between landmarks and the different viewpoints from which sensors observe these landmarks \cite{stumm2016robust}.
Kong \textit{et al.} \cite{kong2020semantic} constructed the semantic graph which abstracts object instances and their relative position.
LOCUS \cite{vidanapathirana2021locus} employed the spatio-temporal higher-order pooling graphs to merge features including appearance, topology, and temporal links for a unified scene depiction.
The topological semantic graph is designed to enable the goal-directed exploration \cite{kim2023topological}.
The Hydra \cite{hughes2022hydra} system construct 3D scene graph to represent places with a hierarchical graph structure from low-level metric maps to high-level object semantics.
Comparing the similarity of two graphs become the key challenge in PR.
Related solutions including graph kernel formulation \cite{stumm2016robust}, inner product of features from graph neural network \cite{kong2020semantic}, Euclidean distance after feature pooling \cite{vidanapathirana2021locus}, and hierarchical descriptor matching \cite{hughes2022hydra} have been proposed.

\subsubsection{Embeddings}
\label{sec:representation_embedding}
Implicit embeddings differ fundamentally from global descriptors (Section \ref{sec:representation_low_level}): while descriptors are explicitly trained for PR, embeddings are generic latent codes from networks optimized for unrelated tasks.
Despite lacking PR-specific training, these embeddings encode environmental semantics through their native objectives.
Although the 2D grid embeddings of RNR-Map \cite{kwon2023renderable} were trained from image observations via volume rendering techniques, they inherently capture spatial-appearance features usable for PR by cross-correlation.
Their dual utility stems from architectural constraints: the same latent codes enabling rendering also structure information into a format amenable to place matching, proving that task-agnostic embeddings can achieve PR.

Recent works have explored image-language descriptors for PR, leveraging visual-language models (VLM) to link visual and textual information \cite{kassab2024language,lyu2024tell,chen2024scene}.
The CLIP model \cite{radford2021learning} aligns images and text through contrastive learning, projecting both modalities into a shared embedding space where semantically related visual and textual representations exhibit high similarity.
LEXIS \cite{kassab2024language} further integrates CLIP features with topological graph nodes for indoor, room-level PR, employing cosine similarity to gauge distances between image and room text encodings.
These approaches enable the language-based data fusion in PR, demonstrating with enhanced resilience to changes in conditions, viewpoints, and overall generalizability.
Embeddings surpass traditional string comparison \cite{hong2019textplace} by resolving synonym discrepancies despite differing strings. For example, synonyms like ``corridor'' and ``hallway'' yield similar embeddings through semantic alignment, avoiding the brittleness of exact text matches

\subsubsection{Others}
\label{sec:representation_others}
Several novel representations also present potential for PR.
For instance, researchers have used view synthesis methods \cite{torii201524,chen2024neural,cheng2025logs} to increase the density of database images.
Qi \textit{et al.} \cite{qi2024gspr} utilized GS to store multi-modal data (including images and point clouds), eliminating the need for separate feature extraction modules for different sensors.
Moreover, Brachmann \textit{et al.} \cite{brachmann2023accelerated} developed a regression network that implicitly learns place representation by encoding the whole scene as parameters of a neural network, offering advantages in storage and privacy preservation.
This method is categorized as scene coordinate regression \cite{shotton2013scene,liu2025gscpr}: regressing 3D-2D correspondences from the query image to the global scene.

\subsection{Summary}
Place representation methodologies have evolved significantly.
Early approaches relied on handcrafted features requiring substantial domain expertise.
The field transitioned to pre-trained neural networks for feature extraction, enabling end-to-end solutions like NetVLAD tailored for PR. High-level representations introduced robustness to viewpoint and environmental variations while enabling multimodal fusion. Recent vision transformers (ViTs) and foundation models, trained on large-scale datasets, further advanced zero-shot generalization across domains.
The following section discusses place matching strategies that leverage these representations, addressing practical challenges in robotic navigation.

\begin{table*}[t]
  \centering
  \vspace{-0.3cm}
  \caption{Representative solutions to specific PR challenges.}
  \vspace{-0.3cm}
  \renewcommand\arraystretch{1.05}
  \begin{tabular}[t]{p{1.5cm}|p{16cm}}
    \Xhline{0.03cm}
    \textbf{Challenges}             &
    \textbf{Categories of Solutions}                                                                                                                       \\
    \Xhline{0.03cm}

    Appearance \newline Change      &
    \textbf{Place Modeling}: semantics \cite{peng2021semantic}, domain transfer \cite{yin2023isimloc}, event cameras \cite{lee2021eventvlad,hou2023fe} \newline
    \textbf{Place Matching with Sequences}: sequence matching \cite{Milford2012SeqSLAM}, dynamic time warping \cite{lu2021sta}, sequential descriptor \cite{Garg2021SeqNet}
    \\
    \Xhline{0.01cm}

    Viewpoint \newline Difference   &
    \textbf{Geometric}: cylinder projection \cite{chen2021overlapnet,xu2021disco}, multi-view projection \cite{yin2021fusionvlad}, rotation-invariant descriptor \cite{komorowski2021minkloc3d} \newline
    \textbf{Appearance}: semantics \cite{garg2018don}, global descriptor \cite{relja2016netvlad,wang2022transvpr,ali2023mixvpr}, multi-scale feature fusion \cite{hausler2021patch} \newline
    \textbf{Others}: hybrid method \cite{yin2023automerge}, omnidirectional sensors \cite{yin2021i3dloc}
    \\
    \Xhline{0.01cm}

    Generalization \newline Ability &
    \textbf{Network Capability}: transformer \cite{wang2022transvpr}, foundation model \cite{keetha2023anyloc,oquab2023dinov2} \newline
    \textbf{Loss Functions}: rotation triplet \cite{Yin2021spherevlad}, angular \cite{loss:barros2021attdlnet}, divergence \cite{yin2021pse}, soft binary-cross entropy \cite{li2022rinet}, large margin cosine \cite{berton2022rethinking} \newline
    \textbf{Incremental Learning}: loss functions \cite{gao2022airloop,knights2022incloud}, HMM \cite{doan2020hm}, dual-memory mechanism \cite{yin2023bioslam} \newline
    \textbf{Other Methods}: multi-modal information \cite{lai2021adafusion, komorowski2021minkloc++, paolicelli2022learning}, domain transfer \cite{yin2023isimloc}
    \\
    \Xhline{0.01cm}

    Efficiency                      &
    \textbf{Optimal Architecture}: efficient backbone \cite{sandler2018mobilenetv2,fan2022svt} \newline
    \textbf{Novel Network Design}: spiking neural network \cite{hines2023vprtempo} \newline
    \textbf{Non-Learning Method}: context encoding \cite{Kim2018scancontext}, planar features \cite{he2016m2dp} \newline
    \textbf{Effient Sequence Matching}: particle filter \cite{Fast_Yang}, approximate world's nearest neighbor \cite{Fast_Siam}, and HMM \cite{Fast_Peter} \\
    \Xhline{0.01cm}

    Uncertainty \newline Estimation &
    \textbf{Employed in PR}: MC Dropout \cite{gal2016dropout}, deep ensembles \cite{lakshminarayanan2017simple},
    probabilistic place embedding \cite{shi2019probabilistic}, self-teaching uncertainty \cite{cai2022stun} \newline
    \textbf{Employed in Other Tasks}: Laplace approximation \cite{yun2023laplace}
    \\

    \Xhline{0.03cm}
  \end{tabular}
  \label{tab:summary_solution_to_challenge}
  \vspace{-0.4cm}
\end{table*}

\section{Recognizing the Right Place Against Challenges}
\label{sec:solution}

As we stated in Section.~\ref{sec:definition_challenges}, the primary challenges for PR can be categorized into five types:
(1) \textbf{Appearance Change},
(2) \textbf{Viewpoint Difference},
(3) \textbf{Generalization Ability},
(4) \textbf{Efficiency}, and
(5) \textbf{Uncertainty Estimation}.
We will investigate them and review existing solutions separately.

\begin{figure}[t]
  \centering
  \subfigure[]{
    \label{fig:solution_appearance_change_place_modeling}
    \centering
    \includegraphics[width=0.235\textwidth]{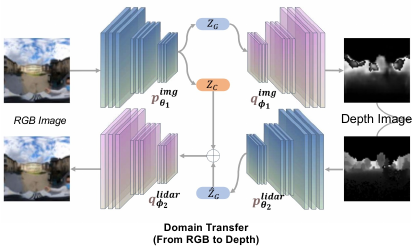}
  }
  \subfigure[]{
    \label{fig:solution_appearance_change_place_matching}
    \centering
    \includegraphics[width=0.20\textwidth]{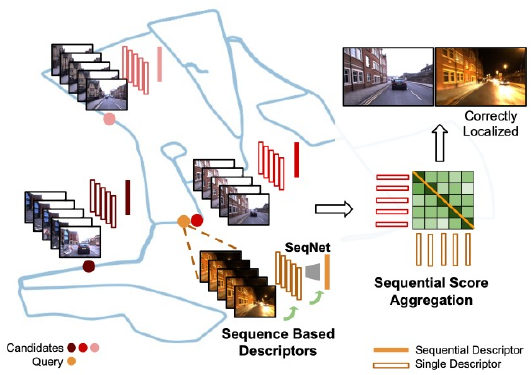}
  }
  \caption{Two typical solutions to appearance change.
    (1) The i3dLoc \cite{yin2021i3dloc} transforms panoramic images from the RGB color domain to the depth image domain, facilitating data association with LiDAR measurements. This method falls under the category of place modeling, as it explicitly models the 3D appearance of environments.
    (2) SeqNet \cite{Garg2021SeqNet} is a sequence-based method, as it employs learned sequential descriptors to compare image sequences.}
  \label{fig:solution_appearance_change}
  \vspace{-0.55cm}
\end{figure}

\subsection{Appearance Change}
\label{sec:appearance_changes}
Two types of changes are commonly presented in PR:
\begin{itemize}
    \item \textit{Conditional changes}, contains the appearance changes caused by environmental conditions, such as illumination, weather, and seasons. This change will mainly affect cameras' observations over time, causing perceptual aliasing and wrong data association.
    \item \textit{Structural changes}, contains the dynamic objects, geometric transformations, and landform changes over short-term or long-term navigation. These changes can be due to natural phenomena, such as seasonal variations and weather conditions, or human activities, including construction and urban development. They mainly affect measurements of range sensors.
\end{itemize}

Two categories of solutions with complementary strengths exist to address appearance changes \cite{Survey:VPR_Deep}:
(1) \textit{place modeling}, which aims to extracting condition-invariant features to represent a place, and
(2) \textit{place mathcing with sequences}, which estimates the place similarity with a sequence of observations.

\subsubsection{Place Modeling}
\label{sec:app_change_place_modeling}
Existing solutions have investigated these strategies:
utilization of additional metric and semantic cues \cite{merrill2019calc2,peng2021semantic}, multi-scale feature fusion \cite{hausler2021patch}, and domain transformation (e.g., transform night-time images into day-time visuals) \cite{yin2019multi}.
CALC2.0 \cite{merrill2019calc2} enhances keypoint extraction by incorporating semantic loss, ensuring the keypoints are semantically contextualized,
while
SRALNet \cite{peng2021semantic} uses semantics as the weight to reinforce local CNN features.
Patch-NetVLAD \cite{hausler2021patch} extends NetVLAD by designing a multi-scale patch feature fusion mechanism, focusing on local details.
Yin \textit{et al.} \cite{yin2019multi} proposed a conditional domain transfer module (CDTM) to transform raw image into simulated image that is condition-invariant.
This solution is also beneficial to cross-modality \cite{yin2021i3dloc} and cross-view \cite{yin2023isimloc} localization.

But several challenges remain in place modeling-based solutions.
For conditional changes, methods have difficulty in generalizing better across a wider range of environmental conditions, especially when training data is limited.
For large structural changes that significantly reshape the spatial layout of a place (e.g., construction sites), systems may fail to detect and adapt to changes during the mission without human intervention.
As a supplement, methods leveraging sequential data can avoid mismatching during single-frame matching.

\subsubsection{Place Matching with Sequences}
\label{sec:app_change_place_matching_seq}

Researchers have explored sequential information for PR, capitalizing on the inherent temporal continuity of robot navigation trajectories.
However, this type of approach fundamentally requires that query and database paths share sequential overlap, \textit{i.e.,} temporal continuity in visited locations.
Milford \textit{et al.} introduced SeqSLAM \cite{Milford2012SeqSLAM}, which replaces single-image matching with sequence alignment: it aggregates similarity scores across consecutive frames using basic normalized image descriptors, significantly improving robustness to environmental variations.

While foundational, SeqSLAM faces two key limitations:
(1) Computational complexity scaling with database size and sequence length, addressed via approximate nearest neighbor search in FastSeqSLAM \cite{Siam2017FastSeqSLAM};
(2) Sensitivity to velocity variations, partially mitigated by Bampis \textit{et al.}'s temporally filtered BoW approach \cite{bampis2018fast}.
To balance recall and efficiency, SeqNet \cite{Garg2021SeqNet} introduced a hierarchical solution using a learned sequential descriptor to generate candidate matches, bypassing exhaustive database searches while suppressing false positives through subsequent sequence verification.
This contrasts with methods such as \cite{Yin2020SeqSphereVLAD,Liu2019SeqLPD} that proposed the coarse-to-fine matching strategies.



\begin{figure}[t]
  \centering
  \includegraphics[width=\linewidth]{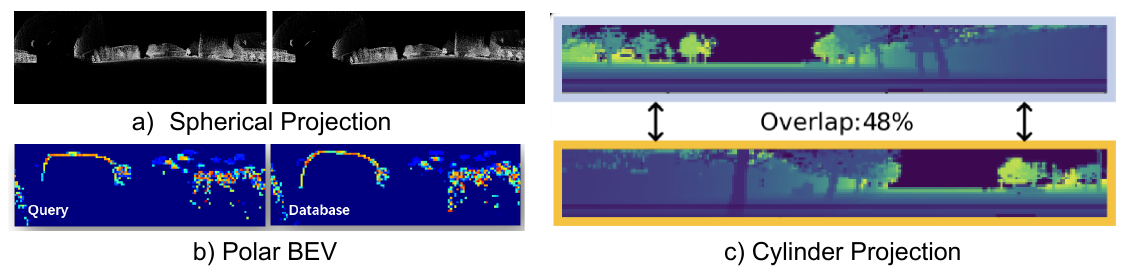}
  \caption{Different projection methods based on geometry can solve the viewpoint difference challenge for LPR \cite{Yin2021spherevlad,Kim2018scancontext,chen2021overlapnet}.}
  \label{fig:solution_viewpoint_difference}
  \vspace{-0.45cm}
\end{figure}

\subsection{Viewpoint Difference}
\label{sec:view_diff}
This challenge is caused by the variation in the perspective from which an environment is observed by sensors.
For instance, observing a building from the front view reveals its full shape, while a top-down view showcases its layout and roof design.
Viewpoint differences also encounter when a robot revisits a location from a different angle or altitude \cite{yin2023isimloc}.
This issue is common in all kinds of sensor modalities and should be handled by PR systems that consist of robust feature extraction and matching modules.

PR solutions to address viewpoint differences can be categorized into three primary groups: geometric, appearance-based, and hybrid approaches, each leveraging different input types.
\textbf{Geometric} methods, predominantly utilized in LPR systems, include innovations like
OverlapNetTransformer \cite{ma2022overlaptransformer}, which transforms yaw differences in point clouds into translational differences on images via cylinder projection, addition to translation-invariant CNN to extract features.
RPR-Net \cite{fan2022rpr} achieves rotation-invariant LPR by utilizing SPRIN \cite{2022Sprin} rotation-invariant local features and geometry constrains which are consistent within different viewpoints.
\textbf{Appearance-based} methods aim to identify visual cues immune to viewpoint shifts.
Semantics-aware PR methods, as proposed by Garg \textit{et al.}, facilitate PR across inverse directions. Techniques such as MixVPR \cite{ali2023mixvpr} utilize global descriptors with attention-weighted patch tokens and isotropic MLP stacks, respectively, to maintain consistent performance despite viewpoint changes.
EigenPlaces~\cite{eigenplaces2023} proposed a novel method to train the network on images from different perspectives.
Patch-NetVLAD \cite{hausler2021patch} focuses on extracting patch-level features for global descriptor computation, enhancing viewpoint invariance through a multiscale patch feature fusion strategy.
\textbf{Hybrid} methods, like AutoMerge \cite{yin2023automerge}, incorporating both point-based (geometry) and projection-based (appearance) feature extraction. This combination addresses the issue caused by translation and orientation disparities, offering a robust framework for PR under varied viewpoints.

Large viewpoint differences may lead to limited overlap between observations, particularly when using pinhole cameras positioned in opposite directions.
Besides above solutions, this challenge can be also mitigated by employing omnidirectional sensors like panorama cameras, LiDARs, and Radars.
Existing studies concentrate on deriving rotation-invariant features and descriptors, employing methods like polar context projection \cite{Kim2018scancontext}, spherical harmonic functions \cite{yin2021i3dloc}, and multi-view fusion \cite{yin2021fusionvlad} to enhance PR under significant viewpoint variations.

\subsection{Generalization Ability}
\label{sec:generalization}
Generalization ability denotes a PR algorithm's capacity to identify locations under environmental conditions absent from its training data.
This capability proves critical when handling domain shifts, such as variations in visual appearance (e.g., day/night cycles), structural layouts (e.g., cross-city topological differences), or perceptual conditions (e.g., perspective versus panoramic imaging).
For instance, an autonomous robot must reliably operate in scenarios ranging from snow-covered rural roads to densely structured urban canyons, despite never encountering these specific configurations during training.
This section introduces four categories of solutions to enhance generalization: domain generalization, domain adaptation, loss functions, and lifelong learning.

\textbf{Domain Generalization} trains models to maintain robustness on unseen target domains by minimizing distributional discrepancies between training and testing environments \cite{wang2022generalizing}.
Building on advancements in place representation such as pre-trained CNNs (e.g., VGG \cite{Feature:VGG}), adaptable architectures like NetVLAD \cite{relja2016netvlad}, attention mechanisms \cite{zhao2022spherevlad2,delf2017}, and Vision Transformers (ViTs) \cite{keetha2023anyloc}, the methods significantly enhance the cross-domain feature transfer.
Another solution is the data manipulation, which enhances training diversity in PR by simulating domain shifts through: geometric transformations (e.g., rotation and resize), image erasing to simulate occlusions, and photometric synthesis of lighting/weather effects \cite{jang2023study}.

\textbf{Domain Adaptation} addresses scenarios where unlabeled target domain data is available during training, enabling model adaptation from a labeled source domain to the target distribution.
Knights \textit{et al.} \cite{knights2023geoadapt} explored test-time adaptation under domain shift between training and test distributions without ground-truth labels.
They proposed GeoAdapt, using geometric consistency to generate pseudo-labels and retrain the model for target domain adaptation.
Semantics that encode high-level human knowledge can enhance PR's generalization \cite{yin2021pse,paolicelli2022learning}. PSE-Match \cite{yin2021pse} separately extracts features from point cloud in different semantics (tree, building, etc.), enabling the descriptors more stable and consistent across different environments.
Visual-LiDAR fusion, as demonstrated in AdaFusion \cite{lai2021adafusion} and MinkLoc++ \cite{komorowski2021minkloc++}, enhances generalization capability in PR, surpassing what a single sensor alone can achieve.


\begin{figure}[t]
  \begin{center}
    \includegraphics[width=\linewidth]{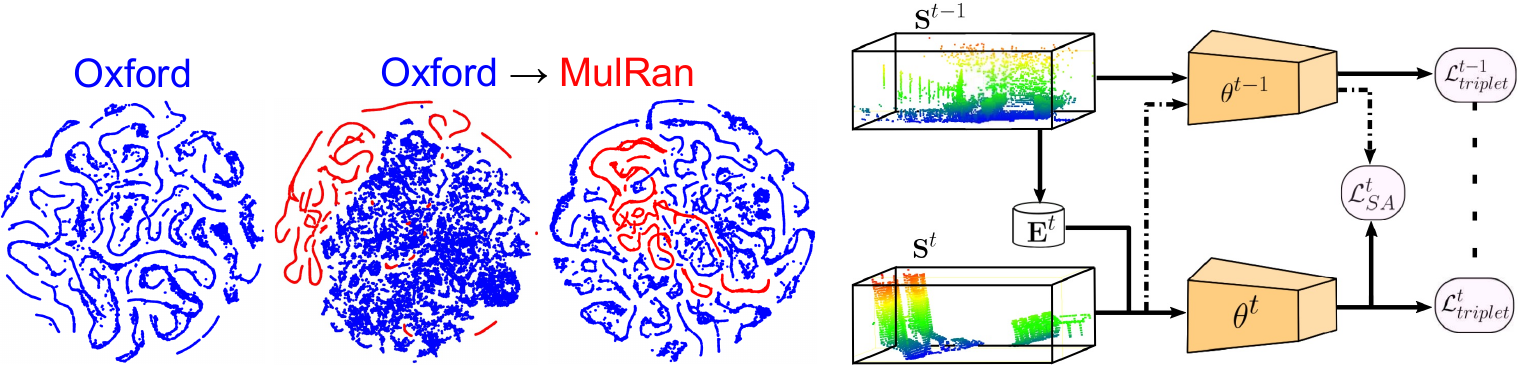}
  \end{center}
  \caption{InCloud \cite{knights2022incloud} introduces an incremental learning approach for LPR, featuring an innovative loss function crafted to maintain the embedding structure during adaptation to new datasets.}
  \label{fig:solution_generalization_ability}
  \vspace{-0.5cm}
\end{figure}

\textbf{Loss Functions} also play a crucial role in training generalized PR models.
Triplet loss, a well-established metric, aims to reduce the distance between query-positive pairs while increasing the separation from negative pairs \cite{relja2016netvlad,uy2018pointnetvlad}.
To enhance orientation invariance, Yin \textit{et al.} \cite{Yin2021spherevlad} proposed the rotation triplet loss.
Angular loss, leveraging cosine similarity, offers robustness against similarity transformations and is effective in handling spatial discrepancies \cite{loss:barros2021attdlnet}.
However, these metrics predominantly concentrate on the relational distances within and between clusters, somewhat overlooking the multifaceted nature of PR challenges.
Alternatively, divergence loss was introduced in \cite{yin2021pse}to target varying semantic structures.
Meanwhile, classification-based strategies like RINet \cite{li2022rinet} formulate PR as classification problem, presenting a soft binary cross entropy loss for the model training.
Berton \textit{et al.} \cite{berton2022rethinking} introduced the Large Margin Cosine Loss to bypass computationally expensive negative sample mining, thereby enabling scalable training on large-scale datasets.

The inherent variability of real-world environments renders exhaustive training data coverage fundamentally unattainable. Rather than pursuing unsustainable model scaling through massive datasets \cite{keetha2023anyloc}, a paradigm shift emerges: \textbf{Lifelong Learning} \cite{wang2024comprehensive} (also termed continual learning) emulates biological learning systems by enabling autonomous agents to incrementally integrate new domain knowledge during operational deployment while maintaining fixed model capacity.
AirLoop \cite{gao2022airloop} proposed two loss fucntions to protect the model from catastrophic forgetting when being adapted to a new domain:
(1) the relational memory-aware synapses loss, which assigns an importance weight to each model parameter, thus regularizing the parameters' adjustments throughout the training process;
(2) the relational knowledge distillation (RKD) loss, designed to preserve the embedding space structure.
In contrast to the RKD loss, InCloud \cite{knights2022incloud} designed a higher-order angular distillation loss.
Fig. \ref{fig:solution_generalization_ability} visulized the key insight of InCloud.
CCL \cite{cui2023ccl} identified the limitation of using triplet loss in InCloud and instead applied contrastive loss to encourage the model to extract more generalizable features. There are approaches, considering the real-world robotic applications, discussed in Section \ref{sec:lifelong_autonomy}.

\subsection{Efficiency}
\label{sec:efficient_robust}
Efficiency in PR involves the system's ability to quickly and accurately recognize previously visited places, which is essential for real-time robotics applications such as loop closure and multi-agent exploration.
Traditional handcrafted methods, including DBoW \cite{galvez2012bags} and the ScanContext series \cite{Kim2018scancontext,wang2020intensity,kim2021scan}, have been widely adopted in real-time SLAM due to their high efficiency.
Conversely, data-driven approaches, though they meet the required performance metrics for large-scale and long-duration navigation tasks, tend to impose substantial computational burdens.
This raises a demand of addressing the efficiency issue.
Overall, the pursuit of efficiency encompasses several dimensions: minimizing time latency, reducing memory usage, and ensuring effective operation on resource-constrained devices without compromising accuracy.

Various strategies have been explored to enhance the efficiency of PR systems, which can be broadly categorized into three primary approaches:
\textbf{Architectures optimized for mobile inference}, focusing on designing systems that are lightweight and capable of running on devices with limited computational resources.
\textbf{Innovative neural network structures}, introducing novel architectures that aim to reduce computational complexity without compromising on the system's ability to accurately recognize places.
\textbf{Accelerated matching with the prior knowledge integration}, leveraging additional information to streamline the recognition process, thus balancing computational demands with recognition accuracy.

Architectural optimization enhances neural network models for greater efficiency \cite{sandler2018mobilenetv2}. MobileNetV2, designed for mobile devices, introduces inverted residual blocks with linear bottlenecks, optimizing both performance and memory efficiency for various vision tasks \cite{sandler2018mobilenetv2}.
FlopplyNet \cite{ferrarini2022binary} proposed the binary neural network with depth reduction and network tunning for VPR.
Oliver \textit{et al.} \cite{grainge2023design} provide an exhaustive analysis of PR efficiency, exploring architectural optimization, pooling methods, descriptor size, and quantization schemes. Their findings suggest that a balance between recall performance and resource consumption is achievable, offering design recommendations for PR systems facing resource constraints.

Researchers have explored the Spiking Neural Network (SNN) \cite{maass1997networks} for PR, leveraging its ability to process information through discrete spikes. This event-driven computation in SNNs, triggered only by significant input changes, drastically reduces energy consumption and computational load, making it ideal for robotics where energy efficiency and real-time processing are paramount.
VPRTempo \cite{hines2023vprtempo} enhances PR efficiency by using temporal coding for spike timing based on pixel intensity, enabling rapid training and querying suitable for resource-limited platforms.
Further, Hussaini \textit{et al.} \cite{hussaini2023applications} introduce three key SNN advancements: modular architecture, ensemble techniques, and sequence matching.

While sequence matching improves localization accuracy, the brute-force approach in SeqSLAM \cite{Milford2012SeqSLAM} remains computationally intensive, though integrating odometry \cite{pepperell2014all} could enhance performance.
Recent efforts to optimize efficiency include particle filters \cite{Fast_Yang}, approximate nearest neighbor search \cite{Fast_Siam}, dimensionality reduction/quantization \cite{garg2020fast}, Hidden Markov Model (HMM) \cite{Fast_Peter} for sequence modeling, and coarse-to-fine strategies \cite{Yin2019MRS_VPR,Yin2020SeqSphereVLAD,Garg2021SeqNet} that eliminate exhaustive database searches through candidate sequence initialization.


\subsection{Uncertainty Estimation}
\label{sec:solution_Uncertainty}
Uncertainty estimation allows PR systems to assess the reliability of their results, highlighting instances where the model's predictions are less certain.
Uncertainty can be used to determine whether the PR systems perform poorly or if the input data are out-of-distribution.
The sources of uncertainty mainly include the sensor noise, models, and environments (e.g., repetitive environments and conditional changes).
Obtaining uncertainty is sometimes equally important to the recognition outcomes due to requirements raised by downstream tasks such as PGO \cite{carlone2016planar}, graph merging \cite{yin2023automerge}, and localization \cite{hu2024paloc}.
PGO typically needs to solve a large optimization problems that involve thousand of variables,
which requires accurate weighting scores and robust outlier rejection to prevent local minima.

\begin{figure}[t]
  \centering
  \subfigure[Method]{
    \label{fig:solution_uncertainty_estimation_stun_method}
    \centering
    \includegraphics[width=0.215\textwidth]{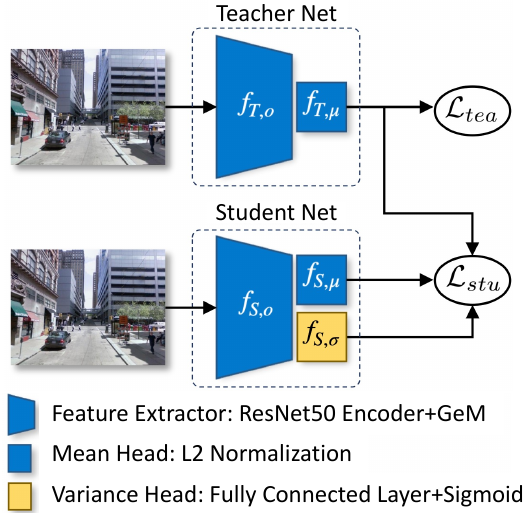}
  }
  \hspace{-0.2cm}
  \subfigure[PR Results]{
    \label{fig:solution_uncertainty_estimation_stun_result}
    \centering
    \includegraphics[width=0.205\textwidth]{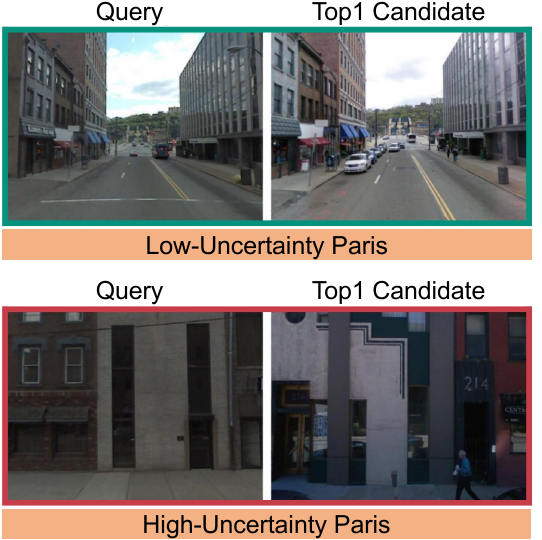}
  }
  \caption{(a) STUN \cite{cai2022stun} proposed the self-teaching uncertainty estimation method by introducing the variance head. (b) PR results which are identified as low-uncertainty and high-uncertainty recognition by STUN, respectively.}
  \label{fig:solution_generalization_ability}
  \vspace{-0.4cm}
\end{figure}

The Bayesian neural network framework is one the pioneering methods in uncertainty estimation for neural networks.
The posterior distribution of network weights can be approximated through Monte-Carlo (MC) Dropout, Deep Ensembles, and Laplace approximation methods \cite{gal2016dropout,kendall2017uncertainties,lakshminarayanan2017simple,yun2023laplace}.
Additionally, some of these methods have been applied to tasks such as semantic segmentation \cite{kendall2017uncertainties} and 3D object detection \cite{yun2023laplace}.
However, they require intensive computational sampling during inference.
Another solution is to integrate evidential theory into neural networks \cite{sensoy2018evidential}. By optimizing the distribution's hyper-parameters, this approach enables precise uncertainty estimation in a single forward pass, thereby eliminating the need for sampling during inference.

Some of these approaches are used to estimate the uncertainty in PR.
Cai \textit{et al.} \cite{cai2022stun} formulated the problem into estimating the distribution of embeddings within the metric space.
They proposed a Student-Teacher Network (STUN), in which a variance-enhanced student network, under the guidance of a pre-trained teacher, refines embedding priors to assess uncertainty at an individual sample level.
Keita \textit{et al.} \cite{mason2023uncertainty} benchmarked existing uncertainty estimation for LPR, including: negative MC-Dropout \cite{gal2016dropout}, Deep Ensembles \cite{lakshminarayanan2017simple}, cosine similarity, Probabilistic Place Embedding (PPE) \cite{shi2019probabilistic}, and STUN \cite{cai2022stun}.
Their findings suggest that, although Ensembles consistently surpass other methods in terms of performance on key LPR datasets, they also demand significant computational resources.

Uncertainty estimation remains a critical and unresolved challenge in PR, characterized by a gap between theory and application.
This complexity arises from several key issues:
(1) Balancing the computational cost with the accuracy of uncertainty estimation for real-time applications.
(2) Avoiding overestimation or underestimation of uncertainty.
(3) Estimation methods for novel foundation model-based PR approaches.
(4) Accurately assessing the uncertainty for sequence-based PR.
Addressing these challenges improve the accuracy and reliability of PR, empowering robots to make informed decisions for subsequent navigation tasks.

\begin{figure*}[t]
  \centering
  \subfigure[Navigation]{
    \label{fig:application_navigation}
    \centering
    \includegraphics[width=0.1665\textwidth]{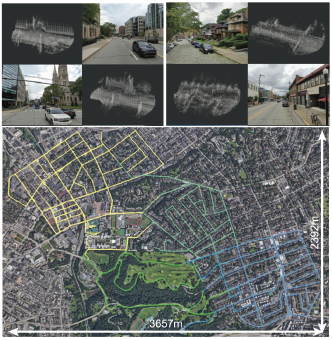}
  }
  \hspace{-0.4cm}
  \subfigure[VRTN]{
    \label{fig:application_navigation}
    \centering
    \includegraphics[width=0.153\textwidth]{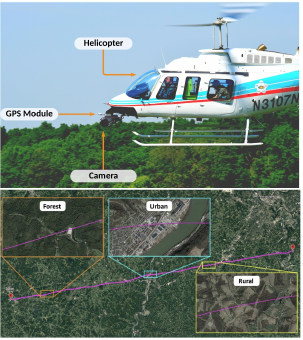}
  }
  \hspace{-0.4cm}
  \subfigure[Multi-Agent]{
    \label{fig:application_navigation}
    \centering
    \includegraphics[width=0.1845\textwidth]{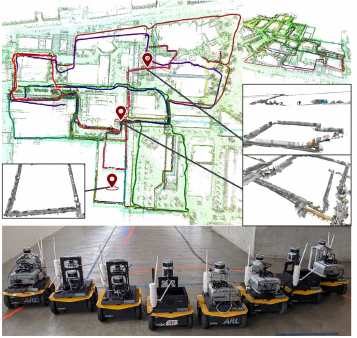}
  }
  \hspace{-0.4cm}
  \subfigure[Lifelong]{
    \label{fig:application_navigation}
    \centering
    \includegraphics[width=0.3285\textwidth]{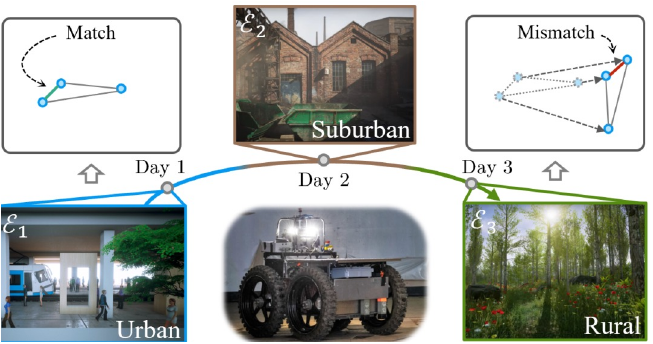}
  }
  \vspace{-0.2cm}
  \caption{PR is beneficial to these roboti applications: (a) long-term and large-scale navigation \cite{yin2022alita}, (b) visual terrain relative navigation \cite{yin2023isimloc}, (c) multi-agent localization and mapping \cite{tian2023resilient}. (d) lifelong autonomy \cite{gao2022airloop}.}
  \label{fig:application}
  \vspace{-0.5cm}
\end{figure*}

\section{Applications \& Trends}
\label{sec:application}

Looking towards the horizon of future applications, multiple potential avenues are currently unfolding withitn the field of PR.
This sectioin delineates four pivotal directions:
(1) \textbf{Long-Term and Large-Scale Navigation} for mobile robots,
(2) \textbf{Visual Terrain Relative Navigation} for aerial robots,
(3) \textbf{Multi-Agent Localization and Mapping}, and
(4) potential pathways to achieve the \textbf{Lifelong Autonomy}.
For each direction, we dive into current status and future opportunities.

\subsection{Long-Term \& Large-Scale Navigation}
\label{sec:app_longterm}
The most in-demand robotic tasks requiring PR is the autonomous navigation,
including applications such as autonomous driving \cite{chen2023end} and subterranean search \cite{tranzatto2022cerberus}.
PR enables robots to obtain their global location with precision up to the topological level in a known environment, despite conditional changes of environments.
This capability is crucial to safe and reliable navigation, as it allows robots to
(1) determine whether this place is visited before,
(2) recorver from failure against the kidnapped problem, and
(3) progressively update and enhance their navigational maps over time.

\subsubsection{Brief Survey}
\textbf{SLAM} uses PR to correct accumulated drift by recognizing previously visited locations despite environmental changes or viewpoint variations. Systems like ORB-SLAM \cite{campos2021orb} and VINS-Mono \cite{qin2018vins} integrate PR modules using the DBoW2 library \cite{galvez2012bags} with BoW-based global descriptors for efficient loop detection, complemented by geometric verification to filter false candidates \cite{yu2024gv}. LiDAR- and Radar-based SLAM \cite{shan2021lvi,adolfsson2023tbv} further improve loop closure without requiring an additional camera.
\textbf{Tracking failure}, where the system fails to correctly associate across adjacent frames, often leads to SLAM collapse and can induce the Kidnapped problem.
Failures may result from issues such as motion blur, occlusion, or hardware disconnections.
Recovery requires relocalizing the robot within the previously built map, posing a more complex PR challenge than simple loop detection, as odometry priors are unavailable.
Chen \textit{et al.} \cite{chen2018submap} addressed this with a submap-based SLAM system that enhances resilience by creating and integrating submaps upon failure, ensuring mapping continuity with DBoW2.
Furthermore, Kuse and Shen \cite{kuse2021learning} optimized \cite{uy2018pointnetvlad} for VPR by introducing an all-pair loss function and decoupled convolutions, which accelerates training convergence and reduces the number of parameters.
This solution was integrated into a stereo-inertial SLAM system, achieving real-time loop closure detection and reliable failure recovery in complex indoor environments.

\textbf{Global localization} broadly encompasses the challenge of determining a global position within a pre-mapped area, especially where GNSS is unreliable.
This context implies a significant initial uncertainty in pose estimation.
Sarlin \textit{et al.} \cite{sarlin2019coarse} proposed a hierarchical localization method using a unified CNN to combine local geometric features with a global descriptor for precise $6$-DoF localization in large environments.
Yin \textit{et al.} \cite{yin2021i3dloc} developed a cross-modality visual localization approach for campus-scale areas, leveraging cross-domain transfer networks to align visual inputs with LiDAR projections for robust long-term navigation.
For large-scale urban roads, Liu \textit{et al.} \cite{liu2019lending} introduced a cross-view matching method integrating orientation and geometric data to improve spatial localization recall rates.

The \textbf{Teach-and-Repeat} (T\&R) framework is an efficient navigation solution for diverse mobile robots \cite{mattamala2021learning}.
Without the requirement of constructing a precise global map, it has achieved great performance in applications such as long-range navigation and planet exploration.
During the \textit{teach} phase, a robot is manually guided along a specific path to generate a topological map, optionally incorporating local metric data.
Subsequently, in the \textit{repeat} phase, the robot autonomously localizes itself within this map to follow the established route, demonstrating an efficient method for traversing pre-determined paths even in changing environments.
Therefore, T\&R systems necessitate a robust PR module to guarantee the precise localization at the topological level.
Chen \textit{et al.} \cite{chen2022self} introduced the sequence matching scheme for enduring T\&R operations.
Mattamala \textit{et al.} \cite{mattamala2021learning} proposed to dynamically choose the most informative camera during the repeat phase under a multi-camera configuration, mitigating the impact of sudden PR variations.
PlaceNav \cite{suomela2023placenav} used PR to limit the number of sub-goal candidates for topological navigation.

\subsubsection{Opportunities}
PR is essential for large-scale and long-term navigation, evolving from mere loop closure detection to a broad spectrum of applications including GNSS-denied global localization, failure recovery, and T\&R navigation.
The field's maturation drives a paradigm shift of robot navigation, establishing spatial awareness as a primary perceptual layer that directly addresses two key challenges:
(1) sustaining large-scale localization accuracy over extended temporal horizons, and
(2) enabling dynamic map reuse/updating mechanisms for persistent environmental modeling, significantly enhancing mission success rates and enabling long-horizontal planning.

Recent innovations such as the hierarchical 3D scen graph proposed by Hughes \textit{et al.} \cite{hughes2022hydra} and the text-image retrieval system \cite{chen2024scene} provide novel PR solutions.
A subsequent innovation involves the joint learning of PR and object detection, as proposed in \cite{zhang2025multiview}, enables scene graph construction by integrating spatial and semantic understanding solely from images.
Additionally, PR also facilitates efficient exploration in tasks such as object/image-directed navigation (e.g., finding a chair in an office) \cite{michel2024object,jiao2025litevloc}.
These investigations are crucial for advancing dynamic and long-term navigation strategies in increasingly complex, ever-changing environments.

\subsection{Visual Terrain Relative Navigation}
\label{sec:vtrn}
VTRN is another representative application of PR by comparing onboard camera images (as the observation)
with pre-acquired geo-referenced satellite imagery (as the database) \cite{VPR:SR_Season}.
VTRN proves especially valuable in GNSS-denied environments by utilizing lightweight cameras and widely accessible satellite data, making it applicable to a broad range of robots, including drones and vehicles.
But challenges including changes in environmental conditions (Section~\ref{sec:appearance_changes}), differences in viewpoints (Section~\ref{sec:view_diff}), and constrained model's generalization ability (Section~\ref{sec:generalization}) are presented.

\subsubsection{Brief Survey}
The temporal disparities between capturing satellite images and onboard sensor images often span years.
Therefore, the conditional changes are mainly attributed to day-night transitions and seasonal variances such as lighting conditions, changes in vegetation, and snow coverage.
Current strategies to mitigate these changes \cite{Intro:ge_uav1} include image transformation and feature matching.


Bhavit \textit{et al.} \cite{Intro:ge_uav1} investigated the use of Normalized Information Distance to align Google Earth (GE) satellite images with unmanned aerial vehicle images,
showing its advantage over traditional photometric error measures in day-night scenarios.
Building on this, an auto-encoder network was introduced to embed raw images, improving robustness to environmental changes and streamlining optimization and storage \cite{Intro:ge_uav2}.
To handle seasonal variations, Anthony \textit{et al.} \cite{VPR:SR_Season} applied a U-Net image transform model for aligning cross-seasonal images, particularly effective at high altitudes where invariant geometric features dominate across seasons.

Most previous studies have overlooked viewpoint variations, such as differences in orientation and altitude.
iSimLoc \cite{yin2023isimloc} leverages NetVLAD for local feature aggragation, improving feature matching using sequential data.
For ground robots, Sarlin \textit{et al.} \cite{sarlin2024snap} introduced neural representations for Ground Elevation images, relying solely on ego-view images and camera poses, generating rich semantics automatically.
Shi \textit{et al.} \cite{shi2023boosting} developed a geometry-enhanced cross-view transformer for view correspondence, and Tang \textit{et al.} \cite{tang2021get} proposed a method that transforms GE images into 2D point collections to align directly with BEV images from LiDAR data.

\subsubsection{Opportunities}
Integrating advanced PR algorithms into VTRN unlocks new possibilities for cutting-edge applications across multiple fields.
In particular, this enhancement improves the reliability of autonomous mobile robot navigation in environments where GNSS signals are blocked \cite{yin2023isimloc}.
PR also benefits planetary exploration \cite{witze2020nasa,ding20222}, providing a consistent global position as complemerty to visual odometry.
Furthermore, aerial-ground coordination introduces new prospects for advanced robotic applications, such as environmental reconstruction and cooperative exploration. Aerial imagery contributes valuable prior knowledge for global path planning and mapping.
By linking aerial with ground images, PR algorithms facilitate an integrated aerial-ground collaboration, evolving the functionality of these systems \cite{miller2024air}.

\subsection{Multi-Agent Localization and Mapping}
\label{sec:app_multi}

Multi-agent systems bring a pivotal shift in addressing complex and dynamic tasks that are beyond the capability of a single agent.
The collaboration among robots significantly increase the efficiency to achieve common goals such as cooperative scene exploration \cite{yan2022mui}.
However, one of major challenges in realizing decentralized multi-agent cooperation is to obtain real-time relative coordinates \textit{w.r.t.} each robot, which become serious in environments characterized by uncertainty and high complexity.
PR methods provide a series of solutions, but as pointed out in Section~\ref{sec:appearance_changes} and Section~\ref{sec:view_diff}, the appearance and viewpoint difference from different agents will cause data association to fail for multi-agent cooperation.

\subsubsection{Brief Survey}

Recent advancements in multi-agent systems (MAS) have introduced diverse PR strategies for collaborative mapping and localization.
Van \textit{et al.} \cite{van2018collaborative} utilized compresses visual features for efficient multi-session mapping on KITTI.
Sasaki \textit{et al.} \cite{sasaki2020map} developed a rover-copter-orbiter system, using satellite images for coordinated localization and optimized rover paths.
Ebadi \textit{et al.} \cite{Merge:LAMP} presented a geometric-based system for unstable environments, employing robust filtering to enhance 3D geometric feature reliability.
Kimera-Multi \cite{kimera_multi} incorporates distributed loop closure detection, while Hydra-Multi \cite{chang2023hydra} enables multi-robot 3D scene graph construction with hierarchical loop closure.
Labbé \textit{et al.} \cite{rtabmap} focused on visual LCD, supporting multi-session mapping without initial trajectory transformations.
These methodologies underscore the evolving landscape of multi-agent localization, setting the groundwork for future cross-disciplinary research \cite{xu2022omni}.
However, challenges remain, particularly in large-scale map merging, where significant perspective and appearance differences pose hurdles. The most recent contribution by Yin~\textit{et al.}~\cite{yin2023automerge} addresses these challenges with a framework for large-scale data association and map merging, extracting viewpoint-invariant place descriptors and filtering unreliable loop closures, marking a significant step forward in the field.


\subsubsection{Opportunities}
The field of MAS is approaching a period of notable developments, with PR contributing significantly to the evolution of autonomous technologies.
Among the most promising avenues is the integration of Neural mapping \cite{brachmann2024scene,kerbl20233d} such as 3D Gaussian Splatting \cite{kerbl20233d} that offers a groundbreaking approach to render photo-realistic from a novel view with sparse and unstructured data.
The PR technique, when applied within systems such as virtual and augmented reality, can enable seamless and immersive interactions among agents such as real-world massively multi-player online games and human-machine interaction.

Furthermore, PR methods facilitate the applications of MAS in GNSS-denied environments such as subterranean scenes \cite{yan2022mui}, factories, and forests, as demonstrated in the drone swarm system \cite{xu2022omni}.
Leveraging the coordination and communication of MAS enhances safety and efficiency in hazardous environments, from deep-sea and space exploration to disaster response, reducing dependence on communication infrastructure. MAS also enables innovative crowd-sourced data collection, as seen in the Tesla FSD system \cite{TeslaAutopilot2023}, where a network of sensor-equipped vehicles generates dynamic urban maps, improving algorithm training and autonomous navigation reliability.

\subsection{Bio-Inspired and Lifelong Autonomy}
\label{sec:lifelong_autonomy}

Recent advancements in space robotics,
as evidenced by NASA's new Mars rover Perseverance~\cite{witze2020nasa} and
CNSA's teleoperated Yutu-2 rover on the Moon~\cite{ding20222},
have underscored the challenges of remote operations and the limits of real-time communication.
These challenges make long-term and real-world autonomy a critical requirement for future robots.
PR serves as a critical component in space and underground exploration, facilitating consistent localization of robots within a global coordinate system.
This capability is essential to long-horizon planning and decision-making.
However, the computational resources available to robots are limited, and the performance of PR models often degrade when faced with new environments.
Thus, developing a lifelong PR system is imperative for sustaining real-world autonomy.
Building on the discussion in Section \ref{sec:generalization}, this section further details how PR enhances the capability of lifelong robotic systems.



\subsubsection{Brief Survey}
Tipaldi \textit{et al.} \cite{tipaldi2013lifelong} introduced a traditional probability-based approach to lifelong localization, leveraging a combination of a particle filter with a HMM to assess dynamic changes in local maps effectively.
Zhao \textit{et al.} \cite{zhao2021general} proposed a novel lifelong LiDAR SLAM framework tailored for extended indoor navigation tasks. This framework primarily employs a multiple-session mapping strategy to construct and refine maps while concurrently optimizing memory usage through a Chow-Liu tree-based method \cite{chow1968approximating}.
Notably, real-world SLAM implementations tend to struggle more significantly with less dynamic objects, such as parked cars, compared to highly dynamic ones, like moving vehicles.
Drawing inspiration from this challenge, Zhu \textit{et al.} \cite{zhu2021lifelongsemi} have developed a semantic mapping-enhanced lifelong localization framework that seamlessly integrates existing object detection techniques to continuously update maps.

Lifelong feature learning is crucial for navigation systems but faces challenges from catastrophic forgetting, the gradual loss of prior knowledge, particularly in dynamic environments.
Most VPR methods operate in short-term or static contexts and struggle with continuous adaptation without performance degradation, as evidenced by VPR benchmarks \cite{VPR_Bench}.
Mactavish \textit{et al.} \cite{mactavish2017visual} addressed this with a visual T\&R framework, enabling online lifelong feature learning through a multi-experience localization mechanism that dynamically matches current observations to past experiences.
Building on this, Chen \textit{et al.} \cite{chen2022self} introduced an experience graph to structurally link temporally disjointed image sequences, facilitating continual data aggregation for network training.

To tackle scalability challenges in lifelong learning systems handling infinite data streams, Doan \textit{et al.} \cite{doan2020hm} proposed a hybrid approach combining the HMM with a two-tiered memory architecture.
By separating active memory (for real-time processing) from passive storage (for long-term retention), their method enables efficient dynamic image transfer, ensuring stable performance with minimal computational overhead.
Yin \textit{et al.} \cite{yin2023bioslam} developed BioSLAM, a lifelong learning framework for VPR.
BioSLAM employs a dual-memory system:
A dynamic memory to rapidly assimilate new observations, and
A static memory to preserve foundational knowledge while integrating novel insights.
This design mitigates catastrophic forgetting and maintains consistent VPR accuracy.
The authors also introduced two evaluation metrics: adaptation efficiency (speed of learning new data) and retention ability (preservation of prior knowledge), to demonstrate BioSLAM's superiority over existing methods in incremental learning scenarios.

\subsubsection{Opportunities}
Although lifelong PR is a relatively nascent area compared to other research direction, it presents significant opportunities, particularly in memory management for long-term navigation tasks.
Motivated by advancements in embodied AI, PR methods diverge from traditional couterpart that depend on pre-trained models using offline databases.
lifelong PR leverages embodied intelligence, enabling robots to engage directly with their environment, accumulate rewards, and learn from ongoing data and experiences.
This capability allows robots to execute more complex tasks and navigate more effectively in dynamic settings, ranging from urban landscapes to unstructured terrains like disaster areas or extraterrestrial environments.
\section{Datasets \& Evaluation}
\label{sec:data_eval}


Open datasets introducing new sensor modalities, challenging scenarios, and diverse challenges are instrumental in driving the development of PR approaches.
To fairly assess the performance of various PR algorithms and identify their limitations, well-designed evaluation metrics are crucial.
In this section, we briefly introduce several public PR datasets, propose a new perspective for evaluation, and discuss open-source libraries relevant to PR.

\begin{table*}[t]
  \centering
  \caption{Typical datasets for evaluating VPR, LPR, and RPR.}
  \renewcommand\arraystretch{1.2}
  \renewcommand\tabcolsep{5pt}
  \begin{tabular}{r|c|c|c|c|c|c}
    \Xhline{0.03cm}
    \textbf{Dataset}
                                                 & \textbf{Scenarios}
                                                 & \makecell[c]{\textbf{Length}}
                                                 & \makecell[c]{\textbf{Sensors}}
                                                 & \makecell[c]{\textbf{Appearance Diversity}}
                                                 & \makecell[c]{\textbf{Viewpoint Diversity}}
                                                 & \makecell[c]{\textbf{Dynamic}}
    \\
    \Xhline{0.03cm}
    Nordland~\cite{sunderhauf2013we}             & Train ride                                  & $748km$  & \text{PinC}                        & Four seasons                & No  & No  \\
    \Xhline{0.01cm}
    Oxford RobotCar~\cite{DATASET:Oxford}        & Urban + Suburban                            & $10km$   & \text{L}, \text{PinC}              & All kinds                   & No  & Yes \\
    \Xhline{0.01cm}
    Mapillary~\cite{warburg2020mapillary}        & Urban + Suburban                            & $4228km$ & \text{PinC}                        & All kinds                   & Yes & Yes \\
    \Xhline{0.01cm}
    KITTI360~\cite{DATASET:KITTI360}             & Urban Street                                & $73.7km$ & \text{L}, \text{PinC}, \text{PanC} & Day-time                    & No  & Yes \\
    \Xhline{0.01cm}
    ALTO~\cite{ivan2022alto}(States)             & Urban+Rural+Nature                          & $50km$   & Top-down \text{PinC}               & Day-time                    & Yes & No  \\
    \Xhline{0.01cm}
    ALITA~\cite{yin2022alita}(City)              & Urban + Terrain                             & $120km$  & \text{L}                           & Day time                    & Yes & Yes \\
    \Xhline{0.01cm}
    ALITA~\cite{yin2022alita}(Campus)            & Campus                                      & $60km$   & \text{L}, \text{PanC}              & Day/Night                   & Yes & Yes \\
    \Xhline{0.01cm}
    Oxford Radar RoboCar \cite{barnes2020oxford} & Urban                                       & $280km$  & \text{L}, \text{R}, \text{PinC}    & Day/Night, Weather, Traffic & Yes & No  \\
    \Xhline{0.03cm}
    \multicolumn{7}{l}{\# \textbf{L}: LiDAR. \textbf{R}: Radar. \textbf{PinC}: Pinhole Camera. \textbf{PanC}: Panoramic Camera.}
  \end{tabular}
  \label{tab:public_datasets}
  \vspace{-0.5cm}
\end{table*}

\subsection{Public Datasets}
\label{sec:datasets}
Table \ref{tab:public_datasets} provides a summary of several commonly utilized PR datasets and highlights the key factors.

\subsubsection{VPR Datasets}

Related datasets predominantly cater to various environmental conditions, including repetitive structures \cite{torii2013visual}, illumination \cite{torii201524,berton2021adaptive}, and seasons \cite{sunderhauf2013we}.
The $24$/$7$ Tokyo \cite{torii201524} and Pitts$30$k \cite{torii2013visual} are two classical VPR datasets for their features in street-view imagery.
The Nordland \cite{sunderhauf2013we}, SVOX \cite{berton2021adaptive}, and Boreas datasets \cite{burnett2023boreas} are designed for cross-seasonal VPR, with the former covering natural environments and the latter two focusing on urban settings.
The NYC-Event-VPR \cite{pan2024nyc} dataset was proposed for event-based VPR in dynamic urban environments.
In the realm of lifelong PR, Warburg \textit{et al.} \cite{warburg2020mapillary} introduced the most extensive VPR dataset to date, covering urban and suburban settings over a span of seven years and documenting various condition changes.
The ALIO dataset \cite{ivan2022alto} presents a comprehensive dataset for the VTRN task, including raw aerial visuals and corresponding satellite imagery.

\subsubsection{LPR Datasets}

LPR shares many benchmark datasets with VPR.
Urban driving datasets like KITTI \cite{DATASET:KITTI} and Oxford RobotCar \cite{DATASET:Oxford} are valuable for evaluating PR in open-road scenarios.
Campus environments are represented by the Newer College dataset \cite{ramezani2020newer}, which provides synchronized LiDAR and stereo-inertial data.
The ALITA dataset \cite{yin2022alita} expands evaluation versatility with $50$ city-scale ($120$ overlapping pairs) and $80$ campus-scale trajectories ($150$ overlapping pairs), supporting tasks ranging from cross-domain recognition to lifelong SLAM.
Emerging natural environment datasets \cite{babin2021large,liu2024botanicgarden,knights2023wild} address challenges like structural ambiguity and dynamic vegetation, driven by growing applications in forestry, agriculture, and subterranean robotics.

\subsubsection{RPR Datasets}
RPR datasets typically feature extreme environments under various weather conditions, including foggy and snowy days, where Radar technology demonstrates significant advantages.
Key datasets such as the Oxford RoboCar Radar \cite{barnes2020oxford}, MulRan \cite{kim2020mulran}, and Boreas \cite{burnett2023boreas} showcase Radar's unique capabilities in challenging visibility conditions.





\begin{figure}[t]
  \centering
  \subfigure[Dataset $1$]{
    \label{fig:star_analysis_day_night}
    \centering
    \includegraphics[width=0.232\textwidth]{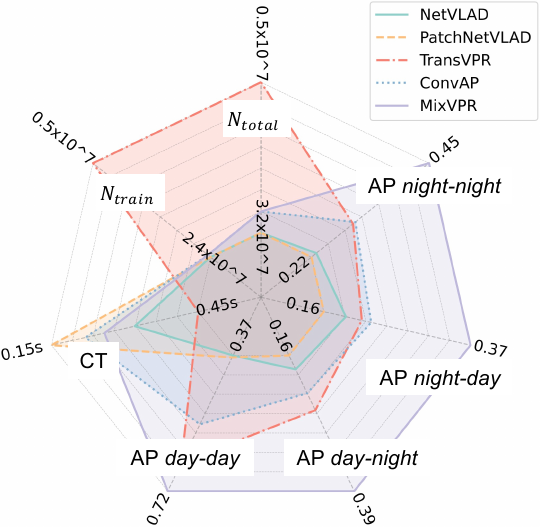}
  }
  \hspace{-0.3cm}
  \subfigure[Dataset $2$]{
    \label{fig:star_analysis_forward_backward}
    \centering
    \includegraphics[width=0.232\textwidth]{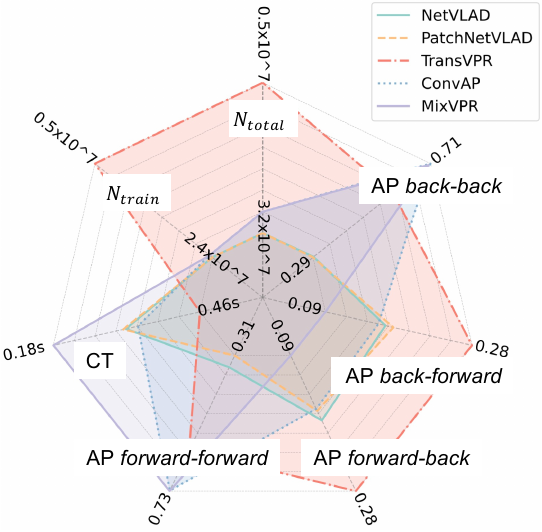}
  }
  \caption{Star-diagram for PR evaluation on two datasets.
    We compare five PR methods in terms of four properties by reporting average precision (AP), number of total and trainable parameters ($N_{\text{total}}$, $N_{\text{train}}$), and average computation time for inference at each frame (CT).
    AP day-day indicates that query and database images are both captured during daytime.}
  \label{fig:star_analysis}
  \vspace{-0.4cm}
\end{figure}

\subsection{New Perspective of Evaluation}
\label{sec:metrics}
As outlined in Section \ref{sec:definition_challenges}, the essential properties of PR encompass condition-invariance, viewpoint-invariance, recognition accuracy, generalization ability, and both training and inference costs.
Utilizing a set of evaluation metrics from VPR-Bench \cite{VPR_Bench}, we propose a comprehensive comparison of methods based on these properties.
We choose the following metrics
\textbf{Average Precision} (AP),
\textbf{Network Parameters} (NP), and
\textbf{Computational Time} (CT).
to illuminate the key characteristics of PR methods:
\begin{itemize}[leftmargin=0.5cm]
    \item \textit{Condition Invariant Property}: AP of PR under different environmental conditions like illumination and weather changes, e.g., comparing night query images against a daytime database.
    \item \textit{Viewpoint Invariant Property}: AP of PR across varying viewpoints, e.g., forward and backward. We consider that environmental conditions and sensors are fixed between the database and query.
    \item \textit{Generalization Ability}: AP of RP in unseen environments after model training, e.g., evaluating how an indoor-trained method performs in urban settings.
    \item \textit{Training and Inference Cost}: Analyzes computational demands, including NP and CT required by a PR algorithm. (extract descriptors)
\end{itemize}


Fig. 12 presents star diagrams comparing five SOTA PR methods (without fine-tuning) using two subsets of the ALITA-campus dataset \cite{yin2022alita}.
Dataset 1 comprises two sequences from the same location captured during daytime and nighttime, was designed to evaluate condition invariance and generalization. The day-night notation denotes a test setting where daytime images form the database and nighttime images serve as queries. Unless stated otherwise, database and query images are randomly selected.
Dataset 2 adopts a similar structure but focuses on viewpoint invariance.
For fair comparison and visualization of method performance, we provide evaluation scripts in this repository\footnote{\url{https://github.com/MetaSLAM/GPRS}}

\section{Conclusion}
\label{sec:conclusion}

The increasing sophistication of mobile robots necessitates navigation systems that sustain indefinite autonomy in expansive, dynamic environments.
PR, which enables robots to identify previously visited locations despite appearance changes and viewpoint variations, has become crucial for robotic autonomy.
This survey outlines significant advancements in PR, detailing its definition, typical representations, strategies for overcoming challenges, and diverse applications. We present a formulation for effective PR, associating it with the need for robotic navigation.

Focusing on the core challenge of ``representing a place,'' we examine the shift from handcrafted features to data-driven methods, benefitting from progress in computer vision and maching learning, especially in neural networks, open-set object detection, and semantic segmentation.
This paradigm shift in high-level representations simplifies PR challenges, enhances model generalization, and creates new opportunities for PR architecture design.
The real-world deployment of PR faces five primary challenges: appearance changes, viewpoint variations, model generalization, resource efficiency, and output uncertainty estimation.
We review key solutions to these challenges, highlighting the research community's gradual pivot from dataset-driven methodologies to systems validated in real-world environments.
PR development has paralleled SLAM advancements, with an increasing number of studies integrating SoTA PR methods to improve navigation systems.
Real-world applications now stretch from large-scale and visual terrain navigation to multi-agent systems, VR/AR, and crowdsourced mapping. The contributions of PR datasets, evaluation metrics, and open-source libraries have been instrumental in advancing the field.

In conclusion, PR holds immense potential for advancing robotic autonomy.
Through this paper and our future efforts, we aim to accelerate progress toward generalized PR, shaping the future of robotic systems and their applications.

\section{Acknowledgement}
The authors thank Jingwen Yu, Jianxing Shi, Xinyi Chen, and Shuyang Zhang for their constructive feedback, and Profs. Michael Milford and Dimitrios Kanoulas for insights on PR definitions and applications. 
We also acknowledge ChatGPT (OpenAI) and DeepSeek (DeepSeek-AI) for text refinement.

\bibliographystyle{IEEEtran}
\bibliography{bible}
\nocite{appsvr2021}
\nocite{vninet2024}
\nocite{gcl2023}
\nocite{covpr2023}
\nocite{nidaloc2024}

\endgroup

\end{document}